\documentclass[twoside,10pt]{article}

\usepackage{amsfonts,epsfig,graphicx}
\usepackage{amsmath,amssymb,amsthm} 
\usepackage[T1]{fontenc} 
\usepackage{epsf} 
\usepackage{graphics} 
\usepackage{amsfonts,amsmath}
\usepackage[ authoryear]{natbib} 
\usepackage{psfrag,xspace}
\usepackage{color,etoolbox}
\usepackage[]{algorithm2e}
\usepackage{bbm}
\usepackage{sidecap}

\usepackage[normalem]{ulem}

\newtheorem{theorem}{Theorem}
\newtheorem{lemma}[theorem]{Lemma} 
\newtheorem{proposition}[theorem]{Proposition} 

\newtheorem{corollary}[theorem]{Corollary}

\usepackage[utf8]{inputenc} 
\usepackage[T1]{fontenc}    
\usepackage{url}            
\usepackage{booktabs}       
\usepackage{amsfonts}       
\usepackage{nicefrac}       
\usepackage{microtype}      

\usepackage{microtype}
\usepackage{graphicx}
\usepackage{float}
\usepackage[export]{adjustbox}
\usepackage{subcaption}
\usepackage{booktabs}
\usepackage{xcolor}

\usepackage{algorithmic}
\usepackage{enumerate}
\usepackage[shortlabels]{enumitem}
\usepackage{bm}
\usepackage{mathtools}

\newcommand{\matsnorm}[2]{|\!|\!| #1 | \! | \!|_{{#2}}}
\newcommand{\frobnorm}[1]{\ensuremath{\matsnorm{#1}{\mbox{\tiny{F}}}}}

\newcommand{\Pbb}{\mathbb{P}}

\newcommand{\Ebb}{\mathbb{E}}

\def\R{\mathbb{R}}
\def\E{\mathbb{E}}
\def\P{\mathbb{P}}

\def\Var{\mathrm{Var}}
\def\half{\frac{1}{2}}

\newcommand{\items}{d}
\newcommand{\popp}{P}
\newcommand{\popq}{Q}
\newcommand{\samples}{k}
\newcommand{\binsamples}{n}
\newcommand{\distra}{p}
\newcommand{\distrb}{q}
\newcommand{\bernoulli}{a}
\newcommand{\weight}{w}
\newcommand{\stat}{T}
\newcommand{\pij}[1]{p_{ij}^{#1}}
\newcommand{\qij}[1]{q_{ij}^{#1}}
\newcommand{\kpij}{\samples^p_{ij}}
\newcommand{\kqij}{\samples^q_{ij}}
\newcommand{\kpji}{\samples^p_{ji}}
\newcommand{\kqji}{\samples^q_{ji}}
\newcommand{\fun}{f}
\newcommand{\fung}{g}
\newcommand{\matx}{X}
\newcommand{\maty}{Y}

\newcommand{\const}{c}
\newcommand{\distance}{\epsilon}
\usepackage{listings}
\newcommand{\nullh}{H_0}
\newcommand{\alt}{H_1}
\newcommand{\perturb}{\delta}
\newcommand{\lbdist}{\eta}
\newcommand{\threshold}{t}
\newcommand{\matex}{M}
\newcommand{\typel}{\alpha}
\newcommand{\typell}{\beta}
\newcommand{\itera}{R}
\newcommand{\iterb}{\gamma}

\newcommand{\pr}{\textnormal{Pr}}
\newcommand{\thetaa}{\theta}
\newcommand{\constfactor}{\nu}

\newcommand{\perturbmat}{\Delta}

\newcommand{\kdist}{{\cal D}}
\newcommand{\kdistinstance}{Z}
\newcommand{\poisson}{\lambda}
\newcommand{\Bin}{\textnormal{Bin}}

\newcommand{\chisq}{\chi^2}
\newcommand{\mset}{\Theta}
\newcommand{\agree}{b_1}
\newcommand{\disagree}{b_2}
\newcommand{\dbytwo}{z}
\newcommand{\ibytwo}{\ell}
\newcommand{\mean}{\mu}
\newcommand{\stddev}{\sigma}
\newcommand{\pone}{p_1}
\newcommand{\pcsize}{\kappa}
\newcommand{\modelclass}{\mathcal{M}}
\newcommand{\risk}{\mathcal{R}}
\newcommand{\length}{m}
\newcommand{\rankdist}[1]{\lambda_{#1}}
\newcommand{\case}{C_1}
\newcommand{\numranks}{N}
\newcommand{\partrank}{\tau}
\newcommand{\subsetm}{\Omega}
\newcommand{\rankbreak}{R}

\makeatletter
\long\def\@makecaption#1#2{
        \vskip 0.8ex
        \setbox\@tempboxa\hbox{\small {\bf #1:} #2}
        \parindent 1.5em  
        \dimen0=\hsize
        \advance\dimen0 by -3em
        \ifdim \wd\@tempboxa >\dimen0
                \hbox to \hsize{
                        \parindent 0em
                        \hfil 
                        \parbox{\dimen0}{\def\baselinestretch{0.96}\small
                                {\bf #1.} #2
                                } 
                        \hfil}
        \else \hbox to \hsize{\hfil \box\@tempboxa \hfil}
        \fi
        }
\makeatother

\begin{document}

\title{Two-Sample Testing on Ranked Preference Data\\ and the Role of Modeling Assumptions}

\author{%
  Charvi Rastogi$^1$, Sivaraman Balakrishnan$^2$, Nihar Shah$^{1,3}$, Aarti Singh$^1$\\
  Machine Learning Department$^1$, Department of Statistics$^2$, Computer Science Department$^3$\\
                    Carnegie Mellon University\\
                    Email: \texttt{\{crastogi@cs, siva@stat, nihars@cs, aarti@cs\}.cmu.edu }
}
\date{}
\maketitle
\begin{abstract}%
A number of applications require two-sample testing on ranked preference data. For instance, in crowdsourcing, there is a long-standing question of whether pairwise-comparison data provided by people is distributed identically to ratings-converted-to-comparisons. Other applications include sports data analysis and peer grading. In this paper, we design two-sample tests for pairwise-comparison data and ranking data. For our two-sample test for pairwise-comparison data, we establish an upper bound on the sample complexity required to correctly test whether the distributions of the two sets of samples are identical. Our test requires essentially no assumptions on the distributions. We then prove complementary lower bounds showing that our results are tight (in the minimax sense) up to constant factors. We investigate the role of modeling assumptions by proving lower bounds for a range of pairwise-comparison models (WST, MST, SST, parameter-based such as BTL and Thurstone). We also provide tests and associated sample complexity bounds for the problem of two-sample testing with partial (or total) ranking data. Furthermore, we empirically evaluate our results via extensive simulations as well as three real-world data sets consisting of pairwise-comparisons and rankings. By applying our two-sample test on real-world pairwise-comparison data, we conclude that ratings and rankings provided by people are indeed distributed differently. 
\end{abstract}

\section{Introduction}

Data in the form of pairwise-comparisons, or more generally partial or total rankings, arises in a wide variety of settings. For instance, when eliciting data from people (say, in crowdsourcing), there is a long-standing debate over the difference between two methods of data collection: asking people to compare pairs of items or asking people to provide numeric scores to the items. A natural question here is whether people implicitly generate pairwise-comparisons using a fundamentally different mechanism than first forming numeric scores and then converting them to a comparison. Thus, we are interested in testing if the data obtained from pairwise-comparisons is distributed identically to if the numeric scores were converted to pairwise-comparisons~\citep{raman2014methods,shah2016estimation}. As another example consider sports and online games, where a match between two players or two teams is a pairwise-comparison between them~\citep{herbrich2007trueskill,hvattum2010using,van2005psychometric}. Again, a natural question that arises here is whether the relative performance of the teams has changed significantly across a certain period of time (e.g., to design an appropriate rating system~\citep{cattelan2013dynamic}). A third example is peer grading where students are asked to compare pairs of homeworks~\citep{shah2013case} or rank a batch of homeworks~\citep{lamon2016whooppee,raman2014methods}. A question of interest here is whether a certain group of students (female/senior/...) grade very differently as compared to another group (male/junior/...)~\citep{shah2018design}. Additionally, consumer preferences as pairwise-comparisons or partial (or total) rankings can be used to investigate whether a certain group (married/old/...) make significantly different choices about purchasing products as opposed to another group (single/young/...)~\citep{cavagnaro2014transitive, regenwetter2011transitivity}. 

Each of the aforementioned problems involves two-sample testing, that is, testing whether the distribution of the data from two populations is identical or not. With this motivation, in this paper we consider the problem of two-sample testing on preference data in the form of pairwise-comparisons and, more generally, partial and total rankings. First, we focus our efforts on preference data in the form of pairwise-comparisons. Specifically, consider a collection of items (e.g., teams in a sports league). The data we consider comprises comparisons between pairs of these items, where the outcome of a comparison involves one of the items beating the other. In the two-sample testing problem, we have access to two sets of such pairwise-comparisons, obtained from two different sources (e.g., the current season in a sports league forming one set of pairwise-comparisons and the previous season forming a second set). The goal is to test whether the underlying distributions (winning probabilities) in the two sets of data are identical or different. Similarly, when the data comprises of partial or total rankings over a collection of items from two different sources, our goal is to test whether the distributions over total rankings for the two sources are identical or not. Specifically, we consider the case where a partial ranking is defined as a total ranking over some subset of the collection of items.

\paragraph{Contributions.} We now outline the contributions of this paper; the theoretical contributions for the pairwise-comparison setting are also summarized in Table~\ref{tab:conversetab}.
\begin{itemize}[leftmargin=*]
    \item First, we present a test for two-sample testing with pairwise-comparison data and associated upper bounds on its minimax sample complexity. Our test makes essentially no assumptions on the outcome probabilities of the pairwise-comparisons. 
    
    \item Second, we prove information-theoretic lower bounds on the critical testing radius for this problem. Our bounds show that our test is minimax optimal for this problem.
    
    \item As a third contribution, we investigate the role of modeling assumptions: What if one could assume one of the popular models (e.g., BTL, Thurstone, parameter-based, SST, MST, WST) for pairwise-comparison outcomes? We show that our test is minimax optimal under WST and MST models. We also provide an information-theoretic lower bound under the SST and parameter-based models. Conditioned on the planted clique hardness conjecture, we prove a computational lower bound for the SST model with a single observation per pair of items, which matches the sample complexity upper bound attained by our test, up to logarithmic factors.
    
    \item Fourth, we conduct experiments on two real-world pairwise-comparison data sets. Our test detects a statistically significant difference between the distributions of directly-elicited pairwise-comparisons and converting numeric scores to comparison data. On the other hand, from the data available for four European football leagues over two seasons, our test does not detect any statistically significant difference between the relative performance of teams across two consecutive seasons. 
    
    \item Finally, we present algorithms for two-sample testing on partial (or total) ranking data for two partial ranking models---namely, the Plackett-Luce model and a more general marginal probability based model. We  provide upper bounds on sample complexity for the test for the Plackett-Luce model controlling both the Type I and Type II error. Moreover, our test for the marginal probability based model controls the Type I error.  We apply our test to a real-world data set on sushi preferences. Our test finds a statistically significant difference in sushi preferences across sections of different demographics based on age, gender and region of residence.
\end{itemize}

\paragraph{Related literature.}
The problem of two-sample testing on ranked preference data is at the intersection of two rich areas of research---two-sample testing and analyzing ranked preference data. The problem of two-sample testing has a long history in statistics, and classical tests include the t-test and Pearson's $\chisq$ test (see for instance \citealp{lehmann2005testing} and references therein). More recently, non-parametric tests \citep{gretton2012kernel,gretton2012optimal,rosenbaum2005exact,kim2020robust,szekely2004testing} have gained popularity but these can perform poorly in structured, high-dimensional settings. 
The minimax perspective on hypothesis testing which we adopt in this work originates in the work of~\cite{ingster94} (and was developed further in~\cite{ingster1997adaptive,ingster03,ingster94}). 
Several recent works have studied the minimax rate for two-sample testing for high-dimensional multinomials \citep{balakrishnan2017hypothesis,balakrishnan2019hypothesis,chan2014optimal,instanceoptimaltestingValiant,Valiant:2011:TSP:2340436.2340454}, and testing for sparsity in regression~\citep{carpentier2018minimax, collier2017minimax}, we build on some of these ideas in our work. We also note the work of~\cite{mania2018kernel} who propose a kernel-based two-sample test for distributions over total rankings.

\begin{table}[tbp]
    \centering
    \begin{tabular}{ |p{2.6cm}|p{3.9cm}|p{4.1cm}|p{4.4cm}| } 
  \hline
 Model $(\modelclass)$ & Upper Bound 
 & Lower Bound 
 & Computational Lower Bound 
 \\ 
 \hline
 Model-free & for $\samples>1$, $\distance_\modelclass^2 \leq \const\dfrac{1}{\samples\items}$ (Thm.~\ref{thm:upperbound}) & $\distance_\modelclass^2  > \const \dfrac{\mathbb{I}(\samples>1)}{\samples\items}  + \dfrac{\mathbb{I}(\samples=1)}{4} $ (Prop.~\ref{remarkkone}) & $\distance_\modelclass^2  > \const \dfrac{\mathbb{I}(\samples>1)}{\samples\items} + \dfrac{\mathbb{I}(\samples=1)}{4}$\\ [7pt]
\hline
WST and MST & $ \distance_\modelclass^2 \leq \const\dfrac{1}{\samples\items}$ & $\distance_\modelclass^2  > \const \dfrac{1}{\samples\items}$ (Thm.~\ref{thm:lbmst}) & $\distance_\modelclass^2  > \const \dfrac{1}{\samples\items}$\\[7pt]
 \hline
 SST & $ \distance_\modelclass^2 \leq \const\dfrac{1}{\samples\items}$ & $\distance_\modelclass^2 > \const\dfrac{1}{\samples\items^{3/2}}$ &for $\samples=1$,  $\distance_\modelclass^2 >\dfrac{\const}{\samples\items(\log\log(\items))^2} $  (Thm.~\ref{thm:polytimelb})\\[7pt]
 \hline
 Parameter-based & $\distance_\modelclass^2 \leq \const\dfrac{1}{\samples\items}$ & $ \distance_\modelclass^2 > \const\dfrac{1}{\samples\items^{3/2}} $ (Thm.~\ref{thm:lbbtl}) & $ \distance_\modelclass^2 > \const\dfrac{1}{\samples\items^{3/2}} $\\[7pt] 
 \hline
\end{tabular}

\caption{\small This table  summarizes our results for two-sample testing of pairwise-comparison data (introduced formally in Equation~\ref{eq:test}), for common pairwise-comparison models. Here, $\items$ denotes the number of items,  and we obtain $\samples$ samples (comparisons) per pair of items from each of the two populations. 
In this work, we provide upper and lower bounds on the critical testing radius $\epsilon_\modelclass$, defined in \eqref{eq:criticalrad}. The upper bound in Theorem~\ref{thm:upperbound} is due to the test in Algorithm~\ref{testalgo} which is computationally efficient. We note that the constant $\const$ varies from result to result.} \label{tab:conversetab}
\end{table}

The analysis of pairwise-comparison data is a rich field of study, dating back at least 90 years to the seminal work of Louis~\cite{thurstone1927law} and subsequently ~\cite{bradley1952rank} and ~\cite{luce1959individual}. Along with this, \cite{Plackett1975TheAO} and \cite{luce1959individual} worked on the now-well-known Plackett-Luce model for partial and total rankings. In the past two decades, motivated by crowdsourcing and other applications, there is significant interest in studying such data in a high-dimensional setting, that is, where the number of items $\items$ is not a fixed constant. A number of papers~\citep[and references therein]{shah2016estimation,chen2015spectral,negahban2012iterative,rajkumar2014statistical,szorenyi2015online, plackettluceguiver, plackettlucemaystre} in this space analyze parameter-based models such as the BTL and the Thurstone models for pairwise-comparison data and the Plackett-Luce model for partial (or total) ranking data. Here the goal is usually to estimate the parameters of the model or the underlying ranking of the items. The papers~\citep[and references therein]{ailon2012active,braverman2008noisy,chatterjee2019estimation,chen2018optimal,falahatgar2017maximum,rajkumar2015ranking} also study ranking from pairwise-comparisons, under some different assumptions. 

Of particular interest is the paper by~\cite{Aldous2017EloRA} which uses the BTL model to make match predictions in sports, and also poses the question of analyzing the average change in the performance of teams over time. While this paper suggests some simple statistics to test for change, designing principled tests is left as an open problem. To this end, we provide a two-sample test without any assumptions and with rigorous guarantees, and also use it subsequently to conduct such a test on real-world data. 

A recent line of work~\citep{heckel2019active,shah2017stochastically,shah15simple} focuses on the role of the modeling assumptions in estimation and ranking from pairwise-comparisons. We study the role of modeling assumptions from the perspective of two-sample testing and prove performance guarantees for some pairwise-comparison models.

\paragraph{Organization.}  The remainder of this paper is organized as follows. In Section~\ref{sec:problemsetting}, we formally describe the problem setup and provide some background on the minimax perspective on hypothesis testing. We also provide a detailed description of the pairwise-comparison models studied in this work. In Section~\ref{sec:mainresults} we present our minimax optimal test for pairwise-comparison data and present the body of our main technical results for the pairwise-comparison setting with brief proof sketches and defer technical aspects of the proofs to Section~\ref{sec:proofs}. Then, in Section~\ref{sec:partial_setup} we extend our results for the two-sample testing problem on partial (or total) ranking data. We describe two partial ranking models and provide testing algorithms and associated sample complexity bounds. The corresponding proofs are in Section~\ref{sec:proofs}. In Section~\ref{sec:experiments}, we present our findings from implementing our testing algorithms on three real-world data sets. Furthermore, we present results of simulations on synthetic data which validate our theoretical findings. We conclude with a discussion in Section~\ref{sec:discussion}. 

\section{Background and problem formulation  for pairwise-comparison setting}
\label{sec:problemsetting}
In this section, we provide a more formal statement of the problem of two-sample testing using pairwise-comparison data along with background on hypothesis testing and the associated definition of risk, and various types of ranking models.

\subsection{Problem statement}
\label{sec:pairwise_problemstatement}
Our focus in this paper is on the two-sample testing problem where the two sets of samples come from two potentially different populations. Here, we describe the model of the data we consider in our work. Specifically, consider a collection of $\items$ items. The two sets of samples comprise outcomes of comparisons between various pairs of these items. In the first set of samples, the outcomes are governed by an unknown matrix $\popp \in [0,1]^{\items \times \items}$. The $(i,j)^\text{th}$ entry of matrix $\popp$ is denoted as $\pij{}$, and any comparison between items $i$ and $j$ results in $i$ beating $j$ with probability $\pij{}$, independent of other outcomes.  We assume there are no ties. Analogously, the second set of samples comprises outcomes of pairwise-comparisons between the $\items$ items governed by a (possibly different) unknown matrix $\popq \in [0,1]^{\items \times \items}$, wherein item $i$ beats item $j$ with probability $\qij{}$, the $(i,j)^\text{th}$ entry of matrix $\popq$. For any pair $(i,j)$ of items, we let $\kpij$ and $\kqij$ denote the number of times a pair of items $(i,j)$ is compared in the first and second set of samples respectively. Let $\matx_{ij}$ denote the number of times item $i \in [\items]$ beats item $j \in [\items]$ in the first set of samples, and let $\maty_{ij}$ denote the analogous quantity in the second set of samples. It follows that $\matx_{ij}$ and $\maty_{ij} $ are Binomial random variables independently distributed as $\matx_{ij} \sim \textnormal{Bin}(\kpij , \pij{})$ and $\maty_{ij} \sim \textnormal{Bin}(\kqij, \qij{})$. We adopt the convention of setting $\matx_{ij} = 0$ when $\kpij=0$, and $\maty_{ij} = 0$ when $\kqij=0$, and $\samples^{\distra}_{ii} = \samples^{\distrb}_{ii} = 0$. \\

\noindent Our results apply to both the symmetric and asymmetric settings of pairwise-comparisons: \\

\noindent \emph{Symmetric setting:} The literature on the analysis of pairwise-comparison data frequently considers a symmetric setting where ``$i$ vs. $j$'' and ``$j$ vs. $i$'' have an identical meaning. Our results apply to this setting, for which we impose the additional constraints that $\distra_{ji} = 1 - \pij{}$ and $\distrb_{ji} = 1 - \qij{}$ for all $(i,j) \in [\items]^2$. In addition, for every $1 \leq i \leq j \leq \items$, we set $\kpji = \kqji=  0$ (and hence $\matx_{ji} = \maty_{ji}=0$), and let $\kpij,\; \kqij,\; \matx_{ij}$ and $\maty_{ij}$ represent the comparisons between the pair of items $(i,j)$ . \\

\noindent\emph{Asymmetric setting:} Our results also apply to an asymmetric setting where ``$i$ vs. $j$'' may have a different meaning as compared to ``$j$ vs. $i$''. For instance, in a setting of sports where ``$i$ vs. $j$'' could indicate $i$ as the home team and $j$ as the visiting team. This setting does not impose the  restrictions described in the symmetric setting above.

\paragraph{Hypothesis test.} 
Consider any class $\modelclass$ of pairwise-comparison probability matrices, and any parameter $\distance > 0$. Then,the goal is to test the hypotheses
\begin{align}
    \begin{split}
        \nullh &: \popp = \popq   \\
        \alt &: \frac{1}{\items}\frobnorm{\popp - \popq} \ge \distance,
    \end{split}
    \label{eq:test}
\end{align}
where $ \popp, \popq \in \modelclass$.

\subsection{Hypothesis testing and risk}
\label{sec:testingrisk}

We now provide a brief background on hypothesis tests and associated terminology. In hypothesis testing, the Type I error is defined as the probability of rejecting the null hypothesis $\nullh$ when the null hypothesis $\nullh$ is actually true, an upper bound on the Type I error is denoted by $\typel$; the Type II error is defined as the probability of failing to reject the null when the alternate hypothesis $\alt$ is actually true, an upper bound on Type II error is denoted by $\typell$. The performance of the testing algorithm is evaluated by measuring its Type I error and its power, which is defined as one minus the Type II error.

 Consider the hypothesis testing problem defined in \eqref{eq:test}. We define a test $\phi$ as $\phi : \{\kpij, \kqij, \matx_{ij}, \maty_{ij}\}_{(i,j)\in [\items]^2} \mapsto \{0,1\}$. Let $\Pbb_0$ and $\Pbb_1$ denote the distribution of the input variables under the null and under the alternate respectively. Here, we assume that the variables $\kpij{}$ and $\kqij{}$ are fixed for all $(i,j)\in [\items^2]$. Let $\modelclass_0$ and $\modelclass_{1}(\distance)$ denote the set of matrix pairs $(\popp, \popq)$ that satisfy the null condition and the alternate condition in \eqref{eq:test} respectively. Then, we define the minimax risk~\citep{ingster94,ingster1997adaptive,ingster03,ingster94} as  
\begin{align}
    \risk_\modelclass = \inf_{\phi} \{\sup_{(\popp, \popq)\in\modelclass_0} \Pbb_0(\phi=1) + \sup_{(\popp, \popq)\in\modelclass_{1}(\distance)} \Pbb_1(\phi=0)\},
    \label{eq:risk}
\end{align}
where the infimum is over all $\{0,1\}$-valued tests $\phi$.  It is common to study the minimax risk via a coarse lens by studying instead the critical radius or the minimax separation. The critical radius is the smallest value $\distance$ for which a hypothesis test has non-trivial power to distinguish the null from the alternate. Formally, we define the critical radius as
\begin{align}
    \distance_\modelclass = \inf\{\distance : \risk_\modelclass\leq 1/3\}.
    \label{eq:criticalrad}
\end{align}
The constant $1/3$ is arbitrary; we could use any specified constant in $(0,1)$. In this paper, we focus on providing tight bounds on the critical radius. 
\subsection{A range of pairwise-comparison models}
\label{sec:pairwise_models}
A model for the pairwise-comparison probabilities is a set of matrices in $[0,1]^{\items \times \items}$. In the context of our problem setting, assuming a model means that the matrices $\popp$ and $\popq$ are guaranteed to be drawn from this set. In this paper, the proposed test and the associated guarantees \underline{do not} make any assumptions on the pairwise-comparison probability matrices $\popp$ and $\popq$. In other words, we allow $\popp$ and $\popq$ to be any arbitrary matrices in $[0,1]^{\items \times \items}$. However, there are a number of models which are popular in the literature on pairwise-comparisons, and we provide a brief overview of them here. We analyze the role of these modeling assumptions in our two-sample testing problem. 
In what follows, we let $\matex \in [0,1]^{\items \times \items}$ denote a generic pairwise-comparison probability matrix, with $\matex_{ij}$ representing the probability that item $i \in [\items]$ beats item $j \in [\items]$. The models impose conditions on the matrix $\matex$. 

\begin{itemize}[leftmargin=*]
\item \emph{Parameter-based models:} 
A parameter-based model is associated with some known, non-decreasing function $\fun : \R \rightarrow [0,1]$ such that $\fun(\thetaa) = 1 - \fun(-\thetaa)~~~\forall\ \thetaa\in\R$. We refer to any such function $\fun$ as being ``valid''. The parameter-based model associated to a given valid function $\fun$ is given by
\begin{align}
    \matex_{ij} = \fun(\weight_i - \weight_j) \quad \textnormal{for all pairs } \;(i,j),
    \label{eq:parametric}
\end{align}
for some unknown vector $\weight \in \R^d$ that represents the notional qualities of the $\items$ items. It is typically assumed that the vector $\weight$ satisfies the conditions $\sum_{i \in [\items]} \weight_i = 0$ and that $\|\weight\|_\infty$ is bounded above by a known constant. 

\begin{itemize}
    \item \emph{Bradley-Terry-Luce (BTL) model:} This is a specific parameter-based model with $\fun(\thetaa)= \dfrac{1}{1+e^{-\thetaa}}$.
    \item \emph{Thurstone model:} This is a specific parameter-based model with $\fun(\thetaa) = \Phi(\thetaa)$, where $\Phi$ is the standard Gaussian CDF.
\end{itemize}
\item \emph{Strong stochastic transitivity (SST): } The model assumes that the set of items $[\items]$ is endowed with an unknown total ordering $\pi$, where $\pi(i) < \pi(j)$ implies that item $i$ is preferred to item $j$. 
A matrix $M \in [0,1]^{\items \times \items}$ is said to follow the SST model if it satisfies the shifted-skew-symmetry condition $\matex_{ij} = 1- \matex_{ji}$ for every pair $i,j \in [\items]$ and the condition
\begin{align}
    \matex_{i\ell} \ge \matex_{j\ell}\;  \textnormal{ for every }\; i,j \in [\items] \textnormal{ such that } \pi(i) < \pi(j) \text{ and for every } \ell \in [\items].
    \label{eq:sst}
\end{align}
\item \emph{Moderate stochastic transitivity (MST):} The model assumes that the set of items $[\items]$ is endowed with an unknown total ordering $\pi$. A matrix $\matex \in [0,1]^{\items \times \items}$ is said to follow the MST model if it satisfies $\matex_{ij} = 1- \matex_{ji}$ for every pair $i,j \in [\items]$ and the condition
\begin{align}
\matex_{i\ell} \ge \min\{\matex_{ij}, \matex_{j\ell}\} \;\textnormal{ for every }\; i,j,\ell \in [\items]\; \textnormal{ such that } \pi(i) < \pi(j)  < \pi(\ell).
\label{eq:mst}
\end{align}
\item \emph{Weak stochastic transitivity (WST):} The model assumes that the set of items $[\items]$ is endowed with an unknown total ordering $\pi$. A matrix $\matex \in [0,1]^{\items \times \items}$ is said to follow the WST model if it satisfies $\matex_{ij} = 1- \matex_{ji}$ for every pair $i,j \in [\items]$ and the condition
\begin{align}
\matex_{i j } \ge \frac{1}{2} \; \textnormal{ for every }\; i,j \in [\items]\; \textnormal{ such that } \pi(i) < \pi(j).
\end{align}
\end{itemize}
\noindent\textbf{Model hierarchy: }There is a hierarchy between these models, that is, \{BTL, Thurstone\} $\subset$ parameter-based $\subset $ SST $\subset $ MST $\subset $ WST $\subset $ model-free
\section{Main results for pairwise-comparison setting}
\label{sec:mainresults}

We now present our main theoretical results for pairwise-comparison data.
\subsection{Test and guarantees}
\label{sec:upperbound}

\RestyleAlgo{boxed}

Our first result provides an algorithm for two-sample testing in the problem~\eqref{eq:test}, and associated upper bounds on its sample complexity. Importantly, we do not make any modeling assumptions on the probability matrices $\popp$ and $\popq$. First we consider a per-pair fixed-design setup in Theorem~\ref{thm:upperbound} where for every pair of items $(i,j)$, the sample sizes  $\kpij{}, \kqij$ are equal to $\samples$. Following that, in Corollary~\ref{rem:upperbound}, we consider a random-design setup wherein for every pair of items $(i,j)$, the sample sizes $\kpij, \kqij$ are drawn i.i.d. from some distribution $\kdist$ supported over non-negative integers. 
\begin{algorithm*}[htbp]

 \textbf{Input}: Samples $\matx_{ij}, \maty_{ij}$ denoting the number of times item $i$ beat item $j$ in the observed $\kpij,\kqij$ pairwise-comparisons from populations denoted by probability matrices $\popp, \popq$ respectively. 
 
\textbf{Test Statistic}:

\begin{align}
\hspace*{-0.6cm}
    \stat = \sum_{i=1}^\items\sum_{j=1}^\items\mathbb{I}_{ij}\dfrac{\kqij(\kqij-1)(\matx_{ij}^2-\matx_{ij}) +\kpij(\kpij-1)(\maty_{ij}^2-\maty_{ij}) - 2(\kpij-1)(\kqij-1)\matx_{ij}\maty_{ij} }{(\kpij-1)(\kqij-1)(\kpij+\kqij)}
    \label{eq:teststat}
\end{align}
where $\mathbb{I}_{ij} = \mathbb{I}(\kpij>1)\times\mathbb{I}(\kqij>1)$. \\
\textbf{Output}: If $\stat \geq  11\items$, where $11\items$ is the threshold, then reject the null. 
 \caption{Two-sample test with pairwise-comparisons for model-free setting}
 \label{testalgo}
\end{algorithm*}

\noindent Our test is presented in Algorithm~\ref{testalgo}. The test statistic~\eqref{eq:teststat} is designed such that it has an expected value of zero under the null and a large expected value under the alternate. The following theorem characterizes the performance of this test, thereby establishing an upper bound on the sample complexity of this two-sample testing problem in a random-design setting.

\begin{theorem}
 Consider the testing problem in \eqref{eq:test} with $\modelclass$ as the class of all pairwise probability matrices. Suppose the number of (per pair) comparisons between the two populations is fixed, $\kpij = \kqij = \samples$~~(for all $i \neq j$ in the asymmetric setting and all $i < j$ in the symmetric setting). 
 There is a constant $\const>0$ such that for any $\distance >0$, if $\samples > 1$ and $\distance^2 \geq  \const\dfrac{1}{\samples\items}$, then the sum of Type I error and Type II error of Algorithm~\ref{testalgo} is at most $\frac{1}{3}$.
 \label{thm:upperbound}
\end{theorem}

\noindent The proof is provided in Section~\ref{sec:proofupperbound}. Theorem~\ref{thm:upperbound} provides a guarantee of correctly distinguishing between the null and the alternate with probability at least $\frac{2}{3}$. The value $\frac{2}{3}$ is closely tied to the specific threshold used in the test above. More generally, for any specified constant $\constfactor \in (0,1)$, the test achieves a probability of error at most $\constfactor$ by setting the threshold as $\items\sqrt{\dfrac{24(2-\constfactor)}{\constfactor}}$, with the same order of sample complexity as in Theorem~\ref{thm:upperbound}. Moreover, if the sample complexity is increased by some factor $\itera$, then running Algorithm~\ref{testalgo} on $\itera$ independent instances of the data and taking the majority answer results in error probability that decreases exponentially with $\itera$ as $(\exp(-2\itera))$, while the sample complexity increases only linearly in $\itera$. One can thus have a very small probability of error of, for instance, $d^{-50}$ with $\samples = \widetilde{O}\left(\dfrac{1}{\items\distance^2}\right)$. Later, in Proposition~\ref{remarkkone}, we show that under the fixed $\samples$ condition, $\kpij{} =  \kqij{}  = \samples$, we have that $\samples>1$ is necessary for our two-sample testing problem. It is also interesting to note that the estimation rate to test the hypotheses in \eqref{eq:test} is $\samples = O\left(\frac{\log(\items)}{\distance^2}\right)$ while the rate for our testing algorithm is $\samples = O\left(\frac{1}{\items\distance^2}\right)$.

 Now, we consider the random-design setup wherein for every pair of items $(i,j)$, the sample sizes $\kpij, \kqij$ are drawn i.i.d. from some distribution $\kdist$ supported over non-negative integers. Let $\mean$ and $ \stddev$ denote the mean and standard deviation of distribution $\kdist$ respectively, and let $\pone \coloneqq  \pr_{\kdistinstance\sim\kdist}(\kdistinstance=1)$. We assume that $\kdist$ has a finite mean and that
\begin{align}
     \mean \geq \const_1\pone; \quad \mean \geq \const_2\stddev,
\label{eq:momentprop}
\end{align}
for some constants $\const_1>1$ and $\const_2>1$.
Many commonly occurring distributions obey these properties, for instance, Binomial distribution, Poisson distribution, geometric distribution and discrete uniform distribution, with appropriately chosen parameters. 
 
 \begin{corollary}
  Consider the testing problem in \eqref{eq:test} with $\modelclass$ as the class of all pairwise probability matrices. Suppose the number of comparisons in the two populations $\kpij, \kqij$ are drawn i.i.d. from some distribution $\kdist$ that satisfies \eqref{eq:momentprop}~~(for all $i \neq j$ in the asymmetric setting and all $i < j$ in the symmetric setting). 
 There is a constant $\const>0$ such that if $\distance^2 \geq \const\max\{\dfrac{1}{\mean\items}, \dfrac{1}{\items^2} \}$, then the sum of Type I error and Type II error of Algorithm~\ref{testalgo} is at most $\frac{1}{3}$.  
 \label{rem:upperbound}
 \end{corollary}
 
\noindent The proof of Corollary~\ref{rem:upperbound} is in Section~\ref{sec:proofremupperbound}. In Corollary~\ref{rem:upperbound}, we see that the even under the random-design setup, our test achieves the same testing rate as in the per-pair fixed-design setup considered in Theorem~\ref{thm:upperbound}, for $\mean \leq \items$. 

We now evaluate the performance of Algorithm~\ref{testalgo} when $\kpij, \kqij$ are drawn i.i.d. from one of the following commonly occurring distributions. Consider any arbitrary matrices $\popp$ and $\popq$. We specialise Corollary~\ref{rem:upperbound} to these distributions by stating the sample complexity that guarantees that the probability of error is at most $\frac{1}{3}$ in the two-sample testing problem \eqref{eq:test}, wherein constant $\const$ may depend on $\const_1, \const_2$ for each distribution. Note that, as in Corollary~\ref{rem:upperbound}, we assume $\distance^2\items^2 \geq \const'$ where $\const'$ is some positive constant
\begin{itemize}[leftmargin=*]
    \item Binomial distribution ($\kpij, \kqij \stackrel{\textnormal{iid}}{\sim} \Bin(\binsamples, \bernoulli)$) : Sufficient condition $\binsamples \geq \const \max\{\dfrac{1}{\bernoulli\items\distance^2}, \dfrac{1}{\bernoulli}\}$.
    \item Poisson distribution ($\kpij, \kqij \stackrel{\textnormal{iid}}{\sim} \textnormal{Poisson}(\poisson)$) : Sufficient condition $\poisson \geq \const\max\{\dfrac{1}{\items\distance^2},1\}$. 
    \item Geometric Distribution ($\kpij,\kqij \stackrel{\textnormal{iid}}{\sim} 
    \textnormal{Geometric}(\bernoulli)$): Sufficient condition $\dfrac{1}{\bernoulli}\geq \const \max\{\dfrac{1}{\items\distance^2},1\}$.
    \item Discrete Uniform Distribution ($\kpij, \kqij \stackrel{\textnormal{iid}}{\sim} \textnormal{Unif}(0, \binsamples)$): Sufficient condition $\binsamples \geq \const\max\{ \dfrac{1}{\items\distance^2}, 1\}$.
\end{itemize}

\noindent Next, we note that a sharper but non-explicit threshold in Algorithm~\ref{testalgo} can be obtained using the permutation test method to control the Type I error. We detail this approach in Algorithm~\ref{permalgo_pairwise}. 

\RestyleAlgo{boxed}
\begin{algorithm*}
\textbf{Input} : Samples $\matx_{ij}, \maty_{ij}$ denoting the number of times item $i$ beat item $j$ in the observed $\kpij,\kqij$ pairwise-comparisons from populations denoted by probability matrices $\popp, \popq$ respectively. Significance level $\typel\in(0,1)$. Iteration count $\iterb$.\\
\textbf{(1)} Compute the test statistic $\stat$ defined in \eqref{eq:teststat}.\\
 \textbf{(2)} For $\ell \leftarrow 1 \text{ to } \iterb $ : \\
 \qquad \textbf{(i)} Repeat this step independently for all $i \neq j$ in the asymmetric setting and for\\ \qquad all $i < j$ in the symmetric setting. Collect the $(\kpij+\kqij)$ samples together and \\ \qquad reassign $\kpij$ of the samples chosen uniformly at random to $\popp$ and the rest to $\popq$.\\ \qquad Compute the new values of  $\matx_{ij}$ and $\maty_{ij}$ based on this reassignment.\\
 \qquad \textbf{(ii)} Using the new values of $\matx_{ij}$ and $\maty_{ij}$, recompute the test statistic in \eqref{eq:teststat}.\\ \qquad Denote the computed test  statistic as $\stat_\ell$.\\
\textbf{Output}  :  Reject the null if $p = \sum_{\ell=1}^\iterb \frac{1}{\iterb}\mathbbm{1}(\stat_\ell - \stat) < \typel$.  
 \caption{Permutation test with pairwise-comparisons for model-free setting.}
 \label{permalgo_pairwise}
\end{algorithm*}

\noindent More generally, the results in Theorem~\ref{thm:upperbound} and Corollary~\ref{rem:upperbound} (and the following converse results in Theorem~\ref{thm:lbmst} and Proposition~\ref{remarkkone}) also apply to the two-sample testing problem of comparing two Bernoulli matrices (or vectors) $\popp$ and $ \popq$, wherein each entry of the matrices (or vectors) is a Bernoulli parameter. In this problem, we want to test whether two Bernoulli matrices are identical or not, and we have access to some observations of some (or all) of the underlying Bernoulli random variables.  
 
 We conclude this section with a proof sketch for Theorem~\ref{thm:upperbound}; the complete proof is provided in Section~\ref{sec:proofremupperbound} and \ref{sec:proofupperbound}.

\paragraph{Proof Sketch for Theorem~\ref{thm:upperbound}.} The test statistic $\stat$ is designed to ensure that $\E_{\nullh}[\stat] = 0$ for any arbitrary pairwise probability matrices $\popp, \popq$ such that $\popp=\popq$, and for any values of  $\{\kpij, \kqij\}_{1\leq i,j\leq \items}$. We lower bound the expected value of $\stat$ under the alternate hypothesis as $\E_{\alt}[\stat] \ge \const\samples\items^2\distance^2$ (Lemma~\ref{lem:expectation}). Next, we show that the variance of $\stat$ is upper bounded under the null by $24\items^2$ and under the alternate by $24\items^2 + 4\samples\items^2\distance^2$ (Lemma~\ref{lem:variance}). These lemmas allow us to choose a suitable threshold value of $11\items$. 
Finally, using Chebyshev's inequality comparing the square of expectation with the variance, we obtain the desired upper bound on the sample complexity with guarantees on both Type I and Type II errors.
\subsection{Converse results and the role of modeling assumptions} 
\label{sec:lowerbound}
In this section we look at the role of modeling assumptions on the pairwise-comparison probability matrices in the two-sample testing problem in \eqref{eq:test}. 
\paragraph{Lower bound for MST, WST, and model-free classes.} Having established an upper bound on the rate of two-sample testing without modeling assumptions on the pairwise-comparison probability matrices $\popp, \popq$, we show matching lower bounds that hold under the MST class. The WST and model-free classes are both supersets of MST, and hence the following  guarantees automatically apply to them as well.

\begin{theorem} 
      Consider the testing problem in \eqref{eq:test} with $\modelclass$ as the class of matrices described by the MST model. Suppose we have $\samples$ comparisons for each pair $(i,j)$ from each population. There exists a constant $\const>0$, such that the critical radius $\distance_\modelclass$ is lower bounded as $\distance_\modelclass^2 >  \dfrac{ \const}{\samples\items}$. 
    \label{thm:lbmst}
\end{theorem}

\noindent The lower bound on the rate matches the rate derived for Algorithm~\ref{testalgo} in Theorem~\ref{thm:upperbound}, thereby establishing the minimax optimality of our algorithm (up to constant factors). The MST class is a subset of the WST model class. This proves that Algorithm~\ref{testalgo} is simultaneously minimax optimal under the MST and WST modeling assumptions in addition to the model-free setting. We provide a proof sketch for Theorem~\ref{thm:lbmst} in Section~\ref{sec:proofsketch}; the complete proof is in Section~\ref{sec:prooflbmst}.
\paragraph{Necessity of $\mean > \pone$.}
Recall that the upper bound derived in Theorem~\ref{thm:upperbound} under the model-free setting holds under the assumption that $\samples >1$ and, similarly, Corollary~\ref{rem:upperbound} holds under the assumption that $\mean \geq \const_1\pone$ with $\const_1>1$, as stated in~\eqref{eq:momentprop}. We now state a negative result for the case $\mean \leq \pone$, which implies that $\kpij,\kqij\leq 1\,\forall\,(i,j)$ under the random-design setup and $\samples \leq 1$ under the per-pair fixed-design setup.
\begin{proposition}
    Consider the testing problem in \eqref{eq:test} with $\modelclass$ as the class of all pairwise probability matrices. Suppose we have at most one comparison for each pair $(i,j)$ from each population~~(for all $i \neq j$ in the asymmetric setting and all $i < j$ in the symmetric setting). Then, for any value of $\distance \leq \frac{1}{2}$, the minimax risk defined in \eqref{eq:risk} is at least $\frac{1}{2}$, thus, $\distance_\modelclass^2 \geq \frac{1}{4}$.
    \label{remarkkone}
\end{proposition}
\noindent We provide some intuition for this result here. If $\kpij=\kqij\leq 1 \,\forall\,(i,j)$, then at best one has access to first order information of each entry of $\popp$ and $\popq$, that is, one has access to only $\Pr(\matx_{ij}=1), \Pr(\maty_{ij}=1), \Pr(\matx_{ij}=1, \maty_{ij}=1) $ for each pair $(i,j)$. This observation allows us to construct a case wherein the null and the alternate cannot be distinguished from each other by any test, due to the inaccessibility of higher order information of the underlying Bernoulli random variables. The complete proof is provided in Section~\ref{sec:impossibility}.

\paragraph{Lower bound for parameter-based class.}
We now prove an information-theoretic lower bound for our two-sample testing problem wherein the probability matrices follow the parameter-based model. 

\begin{theorem} 
      Consider the testing problem in \eqref{eq:test}. Consider any arbitrary non-decreasing function $\fun : \R \rightarrow [0,1]$ such that $\fun(\thetaa) = 1 - \fun(-\thetaa)~~\forall\ \thetaa\in\R$, with $\modelclass$ as the parameter-based class of probability matrices associated to the given function. Suppose we have $\samples$ comparisons for each pair $(i,j)$ from each population. There exists a constant $\const>0$, such that the critical radius $\distance_\modelclass$ is lower bounded as $\distance_\modelclass^2 >  \dfrac{ \const}{\samples\items^{3/2}}$.
    \label{thm:lbbtl}
\end{theorem} 
\noindent  This lower bound also applies to probability matrices in the SST class described in \eqref{eq:sst}. We provide a brief proof sketch in Section~\ref{sec:proofsketch}; the complete proof is in Section~\ref{sec:prooflbbtl}. 
 
 \paragraph{Computational lower bound for SST class.}
  Given the polynomial gap between Theorem~\ref{thm:upperbound} and Theorem~\ref{thm:lbbtl}, it is natural to wonder whether there is another  polynomial-time testing algorithm for testing under the SST and/or parameter-based modeling assumption. We answer this question in the negative, for the SST model and single observation model ($\samples = 1$), conditionally on the average-case hardness of the planted clique problem \citep{Jerrum1992LargeCE, Kucera}. In informal terms, the planted clique conjecture asserts that there is no polynomial-time algorithm that can detect the presence of a planted clique of size $\pcsize = o(\sqrt{\items})$ in an Erd\H os-R\'enyi random graph with $\items$ nodes. We construct SST matrices that are similar to matrices in the planted clique problem and as a direct consequence of the planted clique conjecture, we have the following result.
\begin{theorem}
    Consider the testing problem in \eqref{eq:test} with $\modelclass$ as the class of matrices described by the SST model. 
    Suppose the planted clique conjecture holds. Suppose we have one comparison for each pair $(i,j)$ from each population. Then there exists a constant $\const>0$ such that for polynomial-time testing algorithms the critical radius $\distance_\modelclass$ is lower bounded as $\distance_\modelclass^2 > \dfrac{\const}{\items(\log\log(\items))^2}$.
    \label{thm:polytimelb}
\end{theorem}
\noindent Thus, for $\samples = 1$, the computational lower bound on the testing rate for the SST model  matches the rate derived for Algorithm~\ref{testalgo} (up to logarithmic factors). The proof of Theorem~\ref{thm:polytimelb} is provided in Section~\ref{sec:proofpolytime}. We devote the rest of this section to a sketch of the proofs of Theorem~\ref{thm:lbmst} and Theorem~\ref{thm:lbbtl}.
\subsubsection{Proof sketches for Theorem~\ref{thm:lbmst} and Theorem~\ref{thm:lbbtl}} 
\label{sec:proofsketch}

To prove the information-theoretic lower bound under the different modeling assumptions, we construct a null and alternate belonging to the corresponding class of probability matrices. The bulk of our technical effort is devoted to upper bounding the chi-square divergence between the probability measure under the null and the alternate. We then invoke Le Cam's lower bound for testing to obtain a lower bound on the minimax risk which gives us the information-theoretic lower bound. We now look at the constructions for the two modeling assumptions. 

\noindent \emph{Lower bound construction for MST class} (Section~\ref{sec:prooflbmst}). We construct a null and alternate such that under the null $\popp=\popq=[\half]^{\items\times\items}$ and under the alternate $\popp = [\half]^{\items\times\items}$ and $\popq\in \mset$ with $\frac{1}{\items}\frobnorm{\popp-\popq} = \distance$. For this, we define a parameter $\lbdist \in [0,\half]$ and then define $\mset$ as a set of matrices in which the upper right quadrant has exactly one entry equal to $\half + \lbdist$ in each row and each column and the remaining entries above the diagonal are $\half$. The entries below the diagonal follow from the shifted-skew-symmetry condition. We consider the alternate where $\popq$ is chosen uniformly at random from the set $\mset$ of probability matrices in MST class. 

\noindent \emph{Lower bound construction for parameter-based class} (Section~\ref{sec:prooflbbtl}). The construction is same as the construction given above except we define a different set $\mset$ of probability matrices. According to the parameter-based model, the matrices $\popp$ and $\popq$ depend on the vectors $\weight_\distra \in \R^\items$ and $\weight_\distrb \in \R^\items$ respectively. Now, for simplicity in this sketch, suppose that $\items$ is even.  We set $\weight_\distra = [0, \cdots, 0]$, which fixes $\pij{}= \half\; \forall \;(i,j)$. Consider a collection of vectors each with half the entries as $ \perturb$ and the other half as $- \perturb$, thereby ensuring that $\sum_{i \in [\items]}\weight_i = 0$.  We set $\perturb$ to ensure that each of the probability matrices induced by this collection of vectors obey $\frac{1}{\items}\frobnorm{\popp - \popq}= \distance$. We then consider the setting where $\popq$ is chosen uniformly at random from the set of pairwise-comparison probability matrices induced by the collection of values of $\weight_{\popq}$. 

\section{Two-sample testing with partial or total ranking data }
\label{sec:partial_setup}
In this section, we extend our work from the previous sections to two-sample testing for ranking data. We focus on the two-sample testing problem where the two sets of samples from two potentially different populations comprise of partial or total rankings over various subsets of $\items$ items. Specifically, we consider the case where a partial ranking is defined as a total ranking over some subset of $\items$ items. Let $\rankdist{\popp}$ and $\rankdist{\popq}$ be two unknown probability distributions over the set of all $\items$-length rankings. We observe two sets of partial or total rankings, one set from each of two populations. The partial rankings in the first set are assumed to be drawn i.i.d. according to $\rankdist{\popp}$, and the partial rankings in second set are drawn i.i.d. according to $\rankdist{\popq}$. Each sample obtained is a ranking over a subset of items of size ranging from $2$ to $\items$. Henceforth, we use the term total ranking to specify a ranking over all $\items$ items. We assume there are no ties.

\paragraph{Hypothesis test}  Our goal is to test the hypothesis,
    \begin{align}
    \begin{split}
        \nullh &: \lambda_\popp = \lambda_\popq   \\
        \alt &:  \lambda_\popp \neq \lambda_\popq.
    \end{split}
    \label{eq:test_partial}
\end{align}

\noindent In the sequel, we consider this hypothesis testing problem under certain modeling assumptions on $\lambda_\popp$ and $\lambda_\popq$. 

\subsection{Models}
\label{sec:models}
We now describe two partial ranking models that we analyse subsequently. 

\paragraph{Marginal probability based model}  This is a non-parametric partial ranking model that is entirely specified by the probability distribution over all total rankings, given by $\rankdist{\popp}$ in the first population and $\rankdist{\popq}$ in the second population. 
The distribution $\rankdist{}$ defines the partial ranking model for the corresponding population as follows. Let $S_\items$ denote the set of all total rankings over the $\items$ items. Consider some subset of items $\subsetm \subseteq [\items]$ of size $\length \in \{2,\cdots, \items\}$, and let $\partrank_\subsetm $ be a ranking of the items in this set. Then, we define a set of all total rankings that obey the partial ranking $\partrank_\subsetm$ as 
\begin{align}
    S(\partrank_\subsetm) = \{ \partrank \in S_{\items} \;:\;\partrank(\partrank_\subsetm^{-1}(1))< \partrank(\partrank_\subsetm^{-1}(2)) < \cdots < \partrank(\partrank_\subsetm^{-1}(\length)) \}.
\end{align}
The marginal probability based partial ranking model gives the probability of a partial ranking $\partrank_\subsetm$ as
\begin{align}
    \Pbb(\partrank_\subsetm) = \sum_{\partrank \in S(\partrank_{\subsetm})}\lambda(\partrank),
\end{align}
where $\rankdist{}$ represents $\rankdist{\popp}$ or $\rankdist{\popq}$ for the corresponding population. This model defines the probability of a partial ranking similarly to the non-parametric choice model described in \cite{farias2013nonparametric}. In fact, their choice model defined over sets of size 2 is the same as our model over partial rankings of size 2. 
Our model has the desired property that given a partial ranking over the set $\subsetm$ containing item $i$ and item $j$, the marginal probability that item $i$ is ranked higher than item $j$, denoted by $\Pbb(i \succ j)$, does not depend on other items in set $\subsetm$. Subsequently, the marginal probability is expressed as
\begin{align*}
    \Pbb(i \succ j |  \subsetm ) &= \sum_{\partrank\in S(\partrank_\subsetm), \partrank(i) < \partrank(j) }\lambda(\partrank) \\
    &= \sum_{\partrank\in S_\items , \partrank(i) < \partrank(j)} \lambda (\partrank)
\end{align*}
Now, for the two populations we define the marginal probability of pairwise-comparisons over items $(i,j)$ as $\pij{}$ and $\qij{}$ for all pairs $(i,j)$ with $i < j$. Note that this model has the property that $\pij{} = 1-\distra_{ji}$ and $\qij{} = 1- \distrb_{ji}$ for all $(i,j)$. We also note that the Plackett-Luce model described next is a subset of this model.

\paragraph{Plackett-Luce model} This model introduced by \cite{luce1959individual} and \cite{Plackett1975TheAO} is a commonly used parametric model for ranking data. In this model, each item has a notional quality parameter $\weight_i \in \mathbb{R},\;\forall \; i \in [\items]$. Under the Plackett-Luce model, the partial rankings in each population are generated according to the corresponding underlying quality parameters, namely $\weight^\popp_{i\in [\items]}$ and $ \weight^\popq_{i\in[\items]}$. The weight parameters completely define the probability distribution $\lambda$ over the set of all total rankings. In this model, a partial (or total) ranking $\partrank$ is generated in a sequential manner where each item in a ranking is viewed as chosen from the set of items ranked lower. The probability of choosing an item $i$ from any set $S \subseteq [\items]$ is given by $\frac{\exp(\weight_i)}{\sum_{(i' \in S)}\exp(\weight_{i'})}$. To explain the sequential generation procedure, we show an example here,
\begin{align*}
    \mathbb{P}(i_1 \succ i_2 \succ\cdots \succ i_\ell) = \prod_{j=1}^\ell \dfrac{\exp(\weight_{i_j})}{\sum_{j'=j}^\ell \exp(\weight_{i_{j'}})}. 
\end{align*}
An important property of the Plackett-Luce model is that the marginal probability that item $i$ is ranked higher than item $j$, $\mathbb{P}(i \succ j )$ does not depend on the other items in the ranking, in fact, $\mathbb{P}(i \succ j) = \frac{\exp(\weight_i)}{(\exp(\weight_i)+\exp(\weight_j))}$. For each pair $(i,j)$, we denote the marginal probability $\mathbb{P}(i \succ j )$ corresponding to the parameters  $\weight^\popp_{i\in [\items]}$ as $\pij{}$. Similarly, we denote the marginal probability  $\mathbb{P}(i \succ j )$ corresponding to the parameters  $\weight^\popq_{i\in [\items]}$ as $\qij{}$. These pairwise marginal probabilities $\pij{}$ and $\qij{}$ are collected in pairwise-comparison probability matrices $\popp$ and $\popq$ respectively.

Finally, with this notation, we specialise the hypothesis testing problem in \eqref{eq:test_partial} for the two partial ranking models described above, in terms of the pairwise probability matrices $\popp$ and $\popq$. For any given parameter $\distance>0$, we define the two-sample testing problem as 
    \begin{align}
    \begin{split}
        \nullh &: \popp = \popq   \\
        \alt &: \frac{1}{\items}\frobnorm{\popp - \popq} \geq \distance.
    \end{split}
    \label{eq:test_pl}
\end{align}

\noindent We note that under the Plackett-Luce model, the null condition in \eqref{eq:test_partial} is equivalent to the null condition in \eqref{eq:test_pl}. Moreover, under the Plackett-Luce model, difference in two probability distributions $\lambda_\popp$ and $\lambda_\popq$ is captured by difference in the pairwise probability matrices $\popp$ and $\popq$. Thus, we specialise the alternate condition in \eqref{eq:test_partial} in terms of scaled Frobenius distance between the pairwise probability matrices, $\popp$ and $\popq$, denoted by the parameter $\distance$, to get the alternate condition in \eqref{eq:test_pl}. Furthermore, under the marginal probability based model, the null condition in \eqref{eq:test_partial} implies $\popp=\popq$ whereas the converse is not true. That  is, there exist pairs of probability  distributions over the set of all $\items$-length rankings $\lambda_\popp$ and $\lambda_\popq$, that follow  the marginal probability based model with  $\lambda_\popp \neq \lambda_\popq$, such that their corresponding  pairwise probability matrices $\popp$ and $\popq$ are equal.  Thus, under the marginal probability based model, by conducting a test for the hypothesis testing problem in \eqref{eq:test_pl} that controls the Type I error at level $\typel$, we get control over Type I error at level $\typel$ for the hypothesis testing problem in \eqref{eq:test_partial}. \\

\noindent We are now ready to describe our main results for two-sample testing with partial (or total) ranking data. 
\subsection{Main results}
\label{sec:results_partial}

Our testing algorithms for ranking data build upon the test statistic in Algorithm~\ref{testalgo}. To test for difference in probability distributions $\rankdist{\popp}$ and $\rankdist{\popq}$, we first use a rank breaking method to convert the data into pairwise-comparisons, on which we apply the test statistic in \eqref{eq:teststat}. Given a rank breaking method, denoted by $\rankbreak$, and rankings from the two populations, $S_{\popp_i}$ and $S_{\popq_i}$ for $i\in [\numranks]$, then the rank breaking algorithm yields pairwise-comparison data as $\rankbreak(S_{\popp_{i\in[\numranks]}}) = \{\kpij,  \matx_{ij}\}_{(i,j)\in [\items]^2}$ and, similarly, $\rankbreak(S_{\popq_{i\in[\numranks]}}) = \{\kqij,  \maty_{ij}\}_{(i,j)\in [\items]^2}$. Here, $\kpij{}, \kqij{},\matx_{ij}, \maty_{ij}$ represent the pairwise-comparison data as defined in Section~\ref{sec:pairwise_problemstatement}.
Now, we describe three rank breaking methods that we subsequently use in our testing algorithms, Algorithm~\ref{testalgo_partial} and Algorithm~\ref{permalgo}. 

\begin{enumerate}
    \item \textbf{Random disjoint}: In this method, denoted by $\rankbreak_{\textnormal{R}}$, given a set of $\numranks$ partial (or total) rankings, we randomly break each ranking up into pairwise-comparisons such that no item is in more than one pair. In this method, each $\length$-length ranking yields $\lfloor \frac{\length}{2}\rfloor$ pairwise-comparisons. 
    
    \item \textbf{Deterministic disjoint}: We use this rank breaking method, denoted by $\rankbreak_{\textnormal{D}}$, when we have $\numranks$ total rankings. In this method, we deterministically break each ranking into pairwise-comparisons so that no item is in more than one pair. So, we get $\lfloor \frac{\items}{2}\rfloor $ pairwise-comparisons from each total ranking. First, we want the number of samples to be divisible by $\items$, so we throw away $(\numranks{}\bmod\items)$ rankings chosen randomly. Then arbitrarily without looking at the data, partition the remaining rankings into $\lfloor\frac{\numranks}{\items} \rfloor$ disjoint subsets each containing $\items$ rankings. Within each subset, we convert the $\items$ rankings into $\items\lfloor\frac{\items}{2}\rfloor$ pairwise-comparisons deterministically such that we observe at least one comparison between each pair $(i,j)\in [\items]$ with $i < j$. We keep exactly one pairwise-comparison for each pair in a subset. In this manner, we get to observe exactly $\lfloor\frac{\numranks}{\items} \rfloor$ comparisons between each pair of items. 
    
    \item \textbf{Complete}: In this method, denoted by $\rankbreak_{\textnormal{C}}$, given a set of $\numranks$ partial (or total) rankings, we break each ranking into all possible pairwise-comparisons for that ranking. In this method, each $\length$-length ranking yields $\binom{\length}{2}$ pairwise-comparisons. 
    
\end{enumerate}
Now, equipped with the rank breaking methods, we describe our first result which provides an algorithm (Algorithm~\ref{testalgo_partial}) for the two-sample testing problem in~\eqref{eq:test_pl} for the Plackett-Luce model, and associated upper bounds on its sample complexity. 

\RestyleAlgo{boxed}
\begin{algorithm*}[H]

 \textbf{Input} : Two sets $S_{\popp}$ and $S_{\popq}$ of $\length$-length partial rankings, where $2\leq\length \leq d$. The two sets of partial rankings, $S_\popp$ and $S_\popq$ correspond to pairwise probability matrices $\popp$ and $\popq$ respectively, according to the Plackett-Luce model. Rank breaking method,  $\rankbreak \in \{\rankbreak_{\textnormal{R}}, \rankbreak_{\textnormal{D}}, \rankbreak_{\textnormal{C}}\}$.\\
 \textbf{(1)} Using the rank breaking method get \begin{align*}
     \{\kpij{}, \matx_{ij}\}_{(i,j)\in [\items]^2, i<j} \leftarrow \rankbreak(S_\popp); \quad \{\kqij{}, \maty_{ij}\}_{(i,j)\in [\items]^2, i<j} \leftarrow \rankbreak(S_\popq).
 \end{align*}\\
\textbf{(2)} Execute Algorithm~\ref{testalgo}.
 \caption{Two-sample testing with partial ranking data for Plackett-Luce model.}
 \label{testalgo_partial}
\end{algorithm*}

\noindent We note that both Algorithm~\ref{testalgo_partial} and Algorithm~\ref{permalgo}, defined in this section, can be used with any of the three rank breaking methods described. The subsequent guarantees provided depend on the rank breaking method used, as we see in Theorem~\ref{thm:ell_length} and Theorem~\ref{thm:full_length}. 

\noindent In our results for two-sample testing under the Plackett-Luce modeling assumption, we consider two cases. In the first case, for some $\length \in \{2, \cdots,  \items-1\}$, each sample is a ranking of some $\length$ items chosen uniformly at random from the set of $\items$ items. In the second case, the samples comprise of total rankings, that is, $\length = \items$.
The following two theorems characterize the performance of Algorithm~\ref{testalgo_partial} thereby establishing an upper bound on the sample complexity of the two-sample testing problem defined in \eqref{eq:test_pl}. In these theorems we use the disjoint rank breaking methods so that the pairwise-comparisons created from a ranking are independent. 

\begin{theorem}
          Consider the testing problem in \eqref{eq:test_pl} where pairwise probability matrices $\popp$ and $\popq$ follow the Plackett-Luce model. Suppose we have $\numranks$ samples, where for some $\length \in \{2, \cdots,  \items-1\}$, each sample is a ranking of some $\length$ items chosen uniformly at random from the set of $\items$ items. Then, there are positive constants $\const, \const_0, \const_1$ and $\const_2$ such that if $\numranks \geq \const \dfrac{\items^2\log(\items)}{\length}\lceil \dfrac{\const_0}{\items\distance^2}\rceil$ and $\distance \geq \const_1\items^{-\const_2}$, then Algorithm~\ref{testalgo_partial} with the ``Random disjoint'' rank breaking method will correctly distinguish between $\popp = \popq$ and $\frac{1}{\items}\frobnorm{\popp - \popq} \geq \distance$, with probability at least $\frac{2}{3}$.
          \label{thm:ell_length}
\end{theorem}
\noindent The proof of Theorem~\ref{thm:ell_length} is provided in Section~\ref{sec:proofalgo1}. The lower bound assumption on $\distance$ in Theorem~\ref{thm:ell_length} is to ensure that the sufficient number of pairwise comparisons needed after applying the random disjoint rank breaking algorithm, is not very large. Theorem~\ref{thm:ell_length} is a combined result of the random disjoint rank breaking algorithm and the result in Theorem~\ref{thm:upperbound}. When we have total rankings, Algorithm~\ref{testalgo_partial} with the ``Deterministic disjoint'' rank breaking method yields an improvement in the sample complexity by a logarithmic factor. We state this formally in the following theorem. 
\begin{theorem}
        Consider the testing problem in \eqref{eq:test_pl} where pairwise probability matrices $\popp$ and $\popq$ follow the Plackett-Luce model. Suppose we have $\numranks$ samples of total rankings from each population. Then, there are positive constants $\const, \const_1$ and $ \const_2$ such that if $\numranks \geq 2\items\lceil\dfrac{\const}{\items\distance^2}\rceil$ and $\distance \geq \const_1\items^{-\const_2}$, then Algorithm~\ref{testalgo_partial} with the ``Deterministic disjoint'' rank breaking method will correctly distinguish between $\popp = \popq$ and $\frac{1}{\items}\frobnorm{\popp - \popq} \geq \distance$, with probability at least $\frac{2}{3}$.
          \label{thm:full_length}
\end{theorem}
\noindent The proof of Theorem~\ref{thm:full_length} is provided in Section~\ref{sec:prooffulllength}. These two results provide an upper bound on the sample complexity when using partial (and total) rankings for the two-sample testing problem in \eqref{eq:test_pl} under the Plackett-Luce model. In Theorem~\ref{thm:ell_length} and Theorem~\ref{thm:full_length} the lower bound of $\frac{2}{3}$ on probability of success is tied to the specific threshold used in Algorithm~\ref{testalgo} in the same manner as described for Theorem~\ref{thm:upperbound}. Specifically, for any constant $\constfactor\in(0,1)$, Algorithm~\ref{testalgo_partial} can achieve a probability of error at most $\constfactor$ with the same order of sample complexity as mentioned in Theorem~\ref{thm:ell_length} and Theorem~\ref{thm:full_length}. 

Algorithm~\ref{testalgo_partial} addresses the problem of two-sample testing under the Plackett-Luce model. Now, we provide a permutation test based algorithm for the more general, non-parametric model, namely, marginal probability based model. The permutation test method described in Algorithm~\ref{permalgo} gives a sharper (implicit) threshold than that in Algorithm~\ref{testalgo_partial}. Note that Algorithm~\ref{permalgo} doesn't require any assumptions on the length of the partial-ranking data, the partial-ranking data in each population can be of varying lengths. Moreover, as we will see in Theorem~\ref{thm:permutation_test}, the Type I error guarantee of Algorithm~\ref{permalgo} holds even if the pairwise-comparisons created from the rank breaking method are dependent, hence the guarantee does not depend on the choice of the rank breaking method.

The key difference between the permutation testing algorithm for pairwise-comparison data, described in Section~\ref{sec:upperbound}, and the permutation testing algorithm for partial ranking data, described in Algorithm~\ref{permalgo}, is the shuffling step. In our partial ranking based setup, each ranking sample is obtained independent of all else while the pairwise-comparisons obtained from a rank are not necessarily independent of each other. Hence, in the partial ranking based permutation testing algorithm (Algorithm~\ref{permalgo}), we re-distribute ranking samples between the two populations and not the pairwise-comparisons.\\

\RestyleAlgo{boxed}
\begin{algorithm*}[H]

 \textbf{Input} : Two sets of partial rankings $S_{\popp}$ and $S_{\popq}$ from two populations corresponding to the probability distributions $\rankdist{\popp}$ and $\rankdist{\popq}$. Significance level $\typel \in (0,1)$. Rank breaking method,  $\rankbreak \in \{\rankbreak_{\textnormal{R}}, \rankbreak_{\textnormal{D}}, \rankbreak_{\textnormal{C}}\}$. Iteration count $\iterb$. \\
 \textbf{(1)} Using the rank breaking method get \begin{align*}
     \{\kpij{}, \matx_{ij}\}_{(i,j)\in [\items]^2, i<j} \leftarrow \rankbreak(S_\popp); \quad \{\kqij{}, \maty_{ij}\}_{(i,j)\in [\items]^2, i<j} \leftarrow \rankbreak(S_\popq).
 \end{align*} \\
\textbf{(2)} Compute the test statistic $\stat$ defined in \eqref{eq:teststat}.\\
\textbf{(3)} $\{$Repeat $\iterb$ times$\}$ Put the samples in $S_\popp$ and $S_\popq$ together and reassign the samples at random such that the number of samples assigned to each population is the same as before. Repeat Step 1 and Step 2. Denote the computed test statistic as $\stat_\ell$ for the $\ell^{th}$ iteration.  \\
\textbf{Output}  :  Reject the null if $p = \sum_{\ell=1}^\iterb \frac{1}{\iterb}\mathbbm{1}(\stat_\ell - \stat) < \typel$.  
 \caption{Two-sample testing algorithm with partial ranking data for marginal probability based model.}
 \label{permalgo}
\end{algorithm*}
\vspace{0.4cm}

\noindent We now show that Algorithm~\ref{permalgo} controls the Type I error of the two-sample testing problem in \eqref{eq:test_partial} under the more general, marginal probability based partial ranking model. This result relies on considerably weaker assumptions than Theorem~\ref{thm:ell_length} and Theorem~\ref{thm:full_length}. In particular, we do not assume that each ranking is of the same length. We only assume that the (sub)set of items ranked in each sample from each population is sampled independently from the same distribution. Specifically, let there be any probability distribution over all non-empty subsets of $[\items]$. Then, the set of items ranked in each sample for each population is sampled i.i.d. from this distribution. Moreover, the number of samples from the two populations need not be equal.
\begin{theorem}
          Consider any probability distributions $\rankdist{\popp}$ and $\rankdist{\popq}$ and the two-sample testing problem in \eqref{eq:test_partial}. Suppose we have partial ranking data from each population such that the sets of items ranked in each sample in each population is sampled i.i.d. from any probability distribution over all non-empty subsets of $[\items]$. Suppose the partial ranking data follows the marginal probability based model. Then, for any significance level $\typel \in (0,1)$ the permutation testing method of Algorithm~\ref{permalgo} has Type I error at most $\typel$.  
          \label{thm:permutation_test}
\end{theorem}

\noindent The proof of Theorem~\ref{thm:permutation_test} is provided in Section~\ref{sec:proofalgo2}. Recall that the Plackett-Luce model is a special case of the marginal probability based model, and hence as a direct corollary, the guarantees for Algorithm~\ref{permalgo} established in Theorem~\ref{thm:permutation_test} also apply to the Plackett-Luce model.

\section{Experiments}
\label{sec:experiments}
In this section, we present results from experiments on simulated and real-world data sets, to gain a further understanding of the problem of two-sample testing on pairwise-comparison data. 

\subsection{Pairwise-comparison data}
\label{secc:pairwise_exp}
We now describe real-world experiments and synthetic simulations we conduct for two-sample testing on pairwise-comparison data. In these experiments, we use the test statistic we designed in Algorithm~\ref{testalgo} along with the permutation testing method as described in Algorithm~\ref{permalgo_pairwise} to obtain an implicit value of the threshold and control Type I error.

\subsubsection{Synthetic simulations}
\label{sec:simulations}
\begin{figure}[t!]
  \centering
 \begin{subfigure}[t]{.33\textwidth}
  \centering  
  \includegraphics[width=5.2cm]{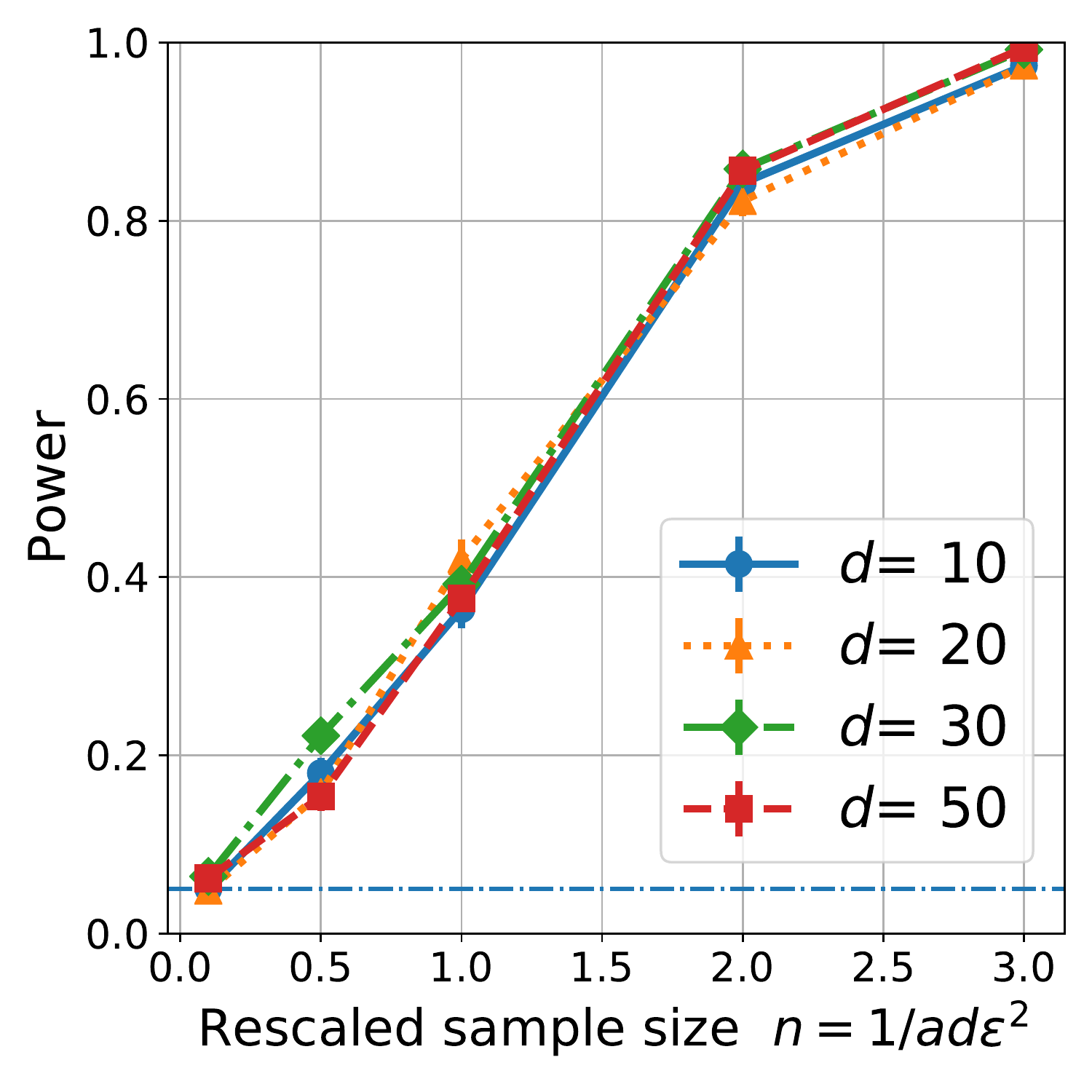}
  \caption{Varying $\items$}
\end{subfigure}%
\begin{subfigure}[t]{.33\textwidth}
  \centering  
  \includegraphics[width=5.2cm]{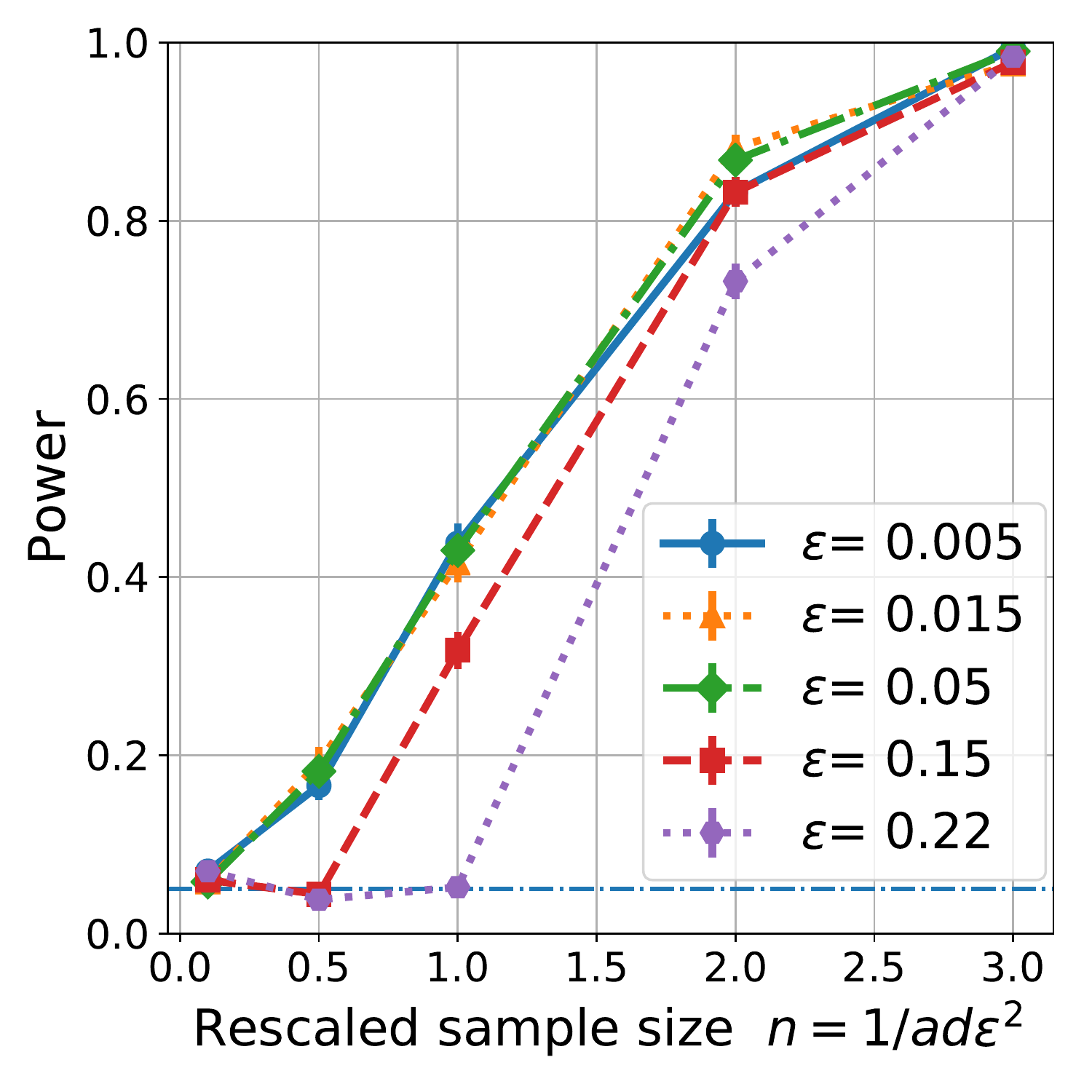}
  \caption{Varying $\distance$}
\end{subfigure}%
\begin{subfigure}[t]{.33\textwidth}
  \centering  
  \includegraphics[width=5.2cm]{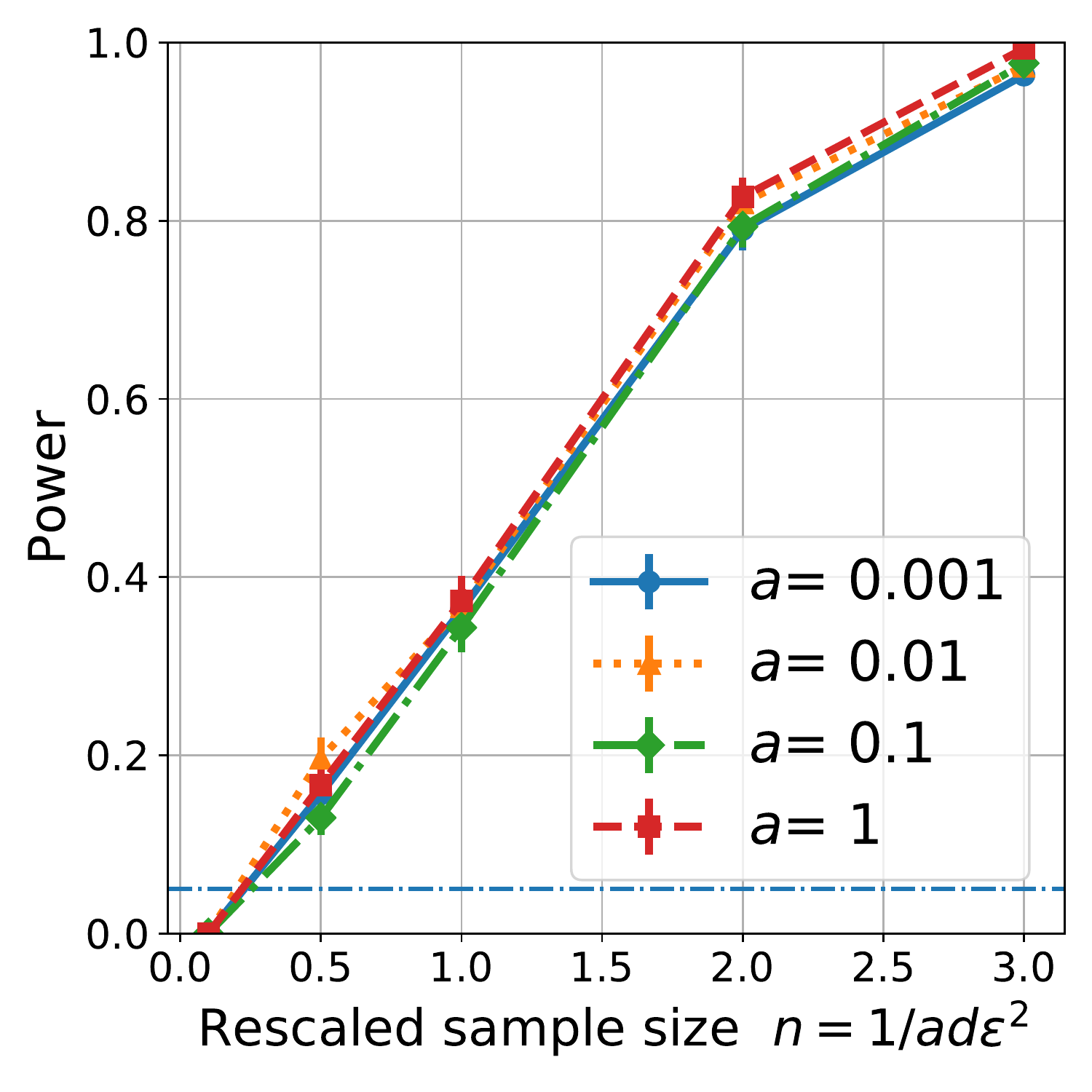}
  \caption{Varying $\bernoulli$}
\end{subfigure}%
    \caption{\small Power of the testing algorithm versus the scaling factor of the sample size parameter $\binsamples = \frac{1}{\bernoulli\items\distance^2}$. We use Algorithm~\ref{permalgo_pairwise} which uses the test statistic in \eqref{eq:teststat} with the permutation testing method. 
    The test is conducted at a significance level of $0.05$ (indicated by the horizontal line at $y =0.05$).  Unless specified otherwise, the parameters are fixed as $\items=20, \distance=0.05, \bernoulli=1$.}
    \label{fig:varyparameter}%
\end{figure}

    We conduct two sets of experiments via synthetic simulations. The first set of experiments empirically evaluates the dependence of the power of our test (Algorithm~\ref{testalgo}) with respect to individual problem parameters. In each of the simulations, we set the significance level to be 0.05. Specifically, given the problem parameters $\binsamples, \bernoulli, \items$ and $\distance$, we consider the random-design setting with $\kpij, \kqij \stackrel{\textnormal{iid}}{\sim} \textnormal{Bin}(\binsamples, \bernoulli)$. We consider the asymmetric and model-free setting, fix $\popp = [\half]^{\items \times \items}$ and set $\popq = \popp + \perturbmat$ where  $\perturbmat$ is sampled uniformly at random from the set of all matrices in $[-\half,\half]^{\items \times \items}$ with $\frac{1}{\items}\frobnorm{\perturbmat} = \distance$. In Figure~\ref{fig:varyparameter}a, b and c, we vary the parameter $\items$, $\distance$ and $\bernoulli$ respectively, keeping the other parameters fixed. Recall that our results in Theorem~\ref{thm:upperbound} and Corollary~\ref{rem:upperbound} predict the sample complexity as $\binsamples = \Theta(\frac{1}{\bernoulli \items\distance^2})$. To test this theoretical prediction, we set the sample size $\binsamples$ (on the x-axis) as $\binsamples = \frac{1}{\bernoulli \items\distance^2}$, and plot the power of the test on the y-axis. Each plot point in Figure~\ref{fig:varyparameter} is obtained by averaging over $400$ iterations of the experiment, and the threshold for the test is obtained by running the permutation test method over $5000$ iterations. Observe that, interestingly in each figure, the curves across all values of the varied parameters nearly coincide, thereby validating the sample complexity predicted by our theoretical results. 

\begin{figure}[t!]
  \centering
  \includegraphics[ height =4.5cm]{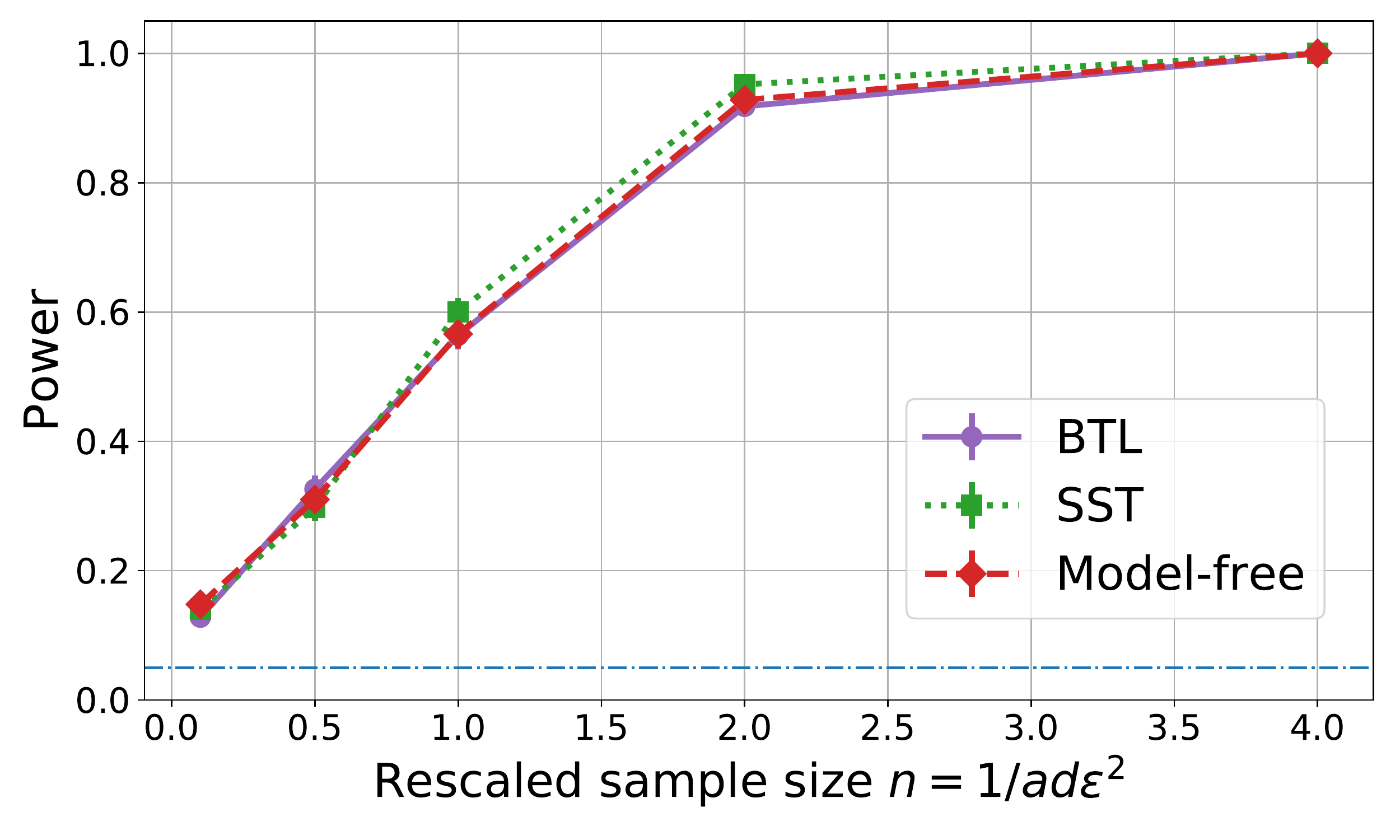}
    \caption{\small Power of our test (Algorithm~\ref{permalgo_pairwise}) under three different models for pairwise-comparisons: BTL, SST and the model-free setting. The parameters of the problem are set as $\items = 20, \distance^2 =0.05, \bernoulli = 1$ and the test is conducted at a significance level of $0.05$ (indicated by the horizontal line at $y =0.05$). }
  \label{fig:model}
\end{figure}

The second set of experiments empirically investigates the role of the underlying pairwise-comparison models in two-sample testing with our test (Algorithm~\ref{testalgo}). We consider the random-design setup in the symmetric setting with $\kpij, \kqij \stackrel{\textnormal{iid}}{\sim} \textnormal{Bin}(\binsamples, \bernoulli)~\forall\; i <j$. We generate the matrices $\popp$ and $\popq$ in three ways: model-free, BTL and SST. In the model-free setting, we generate $\popp$ and $\popq$ in a manner similar to the first set of simulations above, with the additional constraints $\perturbmat_{ji} = 1- \perturbmat_{ij}~\forall \;\;i \leq j$. 
For the BTL and SST models, we  fix $\popp = [\half]^{\items \times \items}$. For the BTL model, we choose $\weight_\distra$ according to the construction in Section~\ref{sec:prooflbbtl} to obtain $\popq$. For the SST model, we set $\popq = \popp + \perturbmat$, where matrix $\perturbmat$ is generated as follows. We generate $\perturbmat $ by arranging $\binom{\items}{2}$ random variables uniformly distributed over $[0,1]$,  in a row-wise decreasing and column-wise increasing order in the upper triangle matrix $(\perturbmat_{ij} = 1- \perturbmat_{ji})$ and normalizing to make $\frac{1}{
\items}\frobnorm{\perturbmat} = \distance$. This construction ensures that  $\perturbmat$ lies in the SST class, and since matrix $\popp$ is a constant matrix, $\popq$ is also guaranteed to  lie in the SST class. The results of the simulations are shown in Figure~\ref{fig:model}. The results show that in the settings simulated, the power of the testing algorithm is identical in all the models considered. This leaves an open question whether there exists a tighter information-theoretic lower bound for the SST and the parameter-based model that matches the upper bound derived for the test in Algorithm~\ref{testalgo} or if there exists a test statistic for these models with a better rate.

\subsubsection{Real-world data}
\label{sec:real_data_experiment}
In this section, we describe the results of our experiments on two real-world data sets. In these experiments, we use Algorithm~\ref{permalgo_pairwise} to obtain a $p$-value for the experiment.
\paragraph{Ordinal versus cardinal}
An important question in the field of crowdsourcing and data-elicitation from people is whether pairwise-comparisons provided by people (ordinal responses) are distributed similarly to if they provide ratings (cardinal responses) which are then  converted to pairwise-comparisons~\citep{shah2016estimation,raman2014methods}. In this section, we use the permutation based two-sample test described in Algorithm~\ref{permalgo_pairwise} to address this question. We use the data set from~\cite{shah2016estimation} comprising six different experiments on the Amazon Mechanical Turk  crowdsourcing platform. In each experiment, workers are asked to either provide ratings for the set of items in that experiment (age for photo given, number of spelling mistakes in a paragraph, distance between two cities, relevance of web-based search results, quality of taglines for a product, frequency of a piano sound clip) or provide pairwise-comparisons. The number of items in each experiment ranged from 10 to 25. For each of the six experiments, there were 100 workers, and each worker was assigned to either the ordinal or the cardinal version of the task uniformly at random. The first set of samples corresponds to the elicited ordinal responses and the second set of samples  are obtained by converting the elicited ratings to ordinal data. We have a total of 2017 ordinal responses  and 1671 cardinal-converted-to-ordinal responses. More details about the data set and experiment are provided in the appendix.  

Using Algorithm~\ref{permalgo_pairwise}, we test for difference in the two resulting distributions for the entire data set  ($\items = 74$).  We observe that \emph{the test rejects the null with a $p$-value of 0.003, thereby concluding a statistically significant difference between the ordinal and the cardinal-converted-to-ordinal data}.

\paragraph{European football leagues} In the second data set, we investigate whether the relative performances of the teams (in four European football leagues: English Premier League, Bundesliga, Ligue 1, La Liga) changed significantly from the 2016-17 season to the 2017-18 season. The leagues are designed such that each pair of teams plays twice in a season (one home, one away game), so we have at most two pairwise-comparisons per pair within a league (we do not consider the games that end in a draw). Each league has 15-17 common teams across two consecutive seasons. This gives a total of 801 and 788 pairwise-comparisons in the 2016-17 and 2017-18 seasons respectively. More details about the experiment are provided in the appendix.

Using the test statistic of Algorithm~\ref{testalgo} with permutation testing, we test for a difference in the two resulting distributions for the entire data set  ($\items = 67$).  We observe that \emph{the test fails to reject the null with a $p$-value of 0.971, that is, it does not recognize any significant difference between the relative performance of the European football teams in 2017-18 season and the 2016-17 season from the data available}. Running the test for each league individually also fails to reject the null. 

\subsection{Partial and total ranking data}
\label{sec:experiments_sushi}
We now describe the experiments we conducted on real-world data for two-sample testing on partial (and total) ranking data. In these experiments, we use the test statistic \eqref{eq:teststat} along with the permutation testing method, as explained in Algorithm~\ref{permalgo}. 

For our experiments, we use the ``Sushi preference data set''~\cite{kamishima2003nantonac}, in which subjects rank different types of sushi according to their preferences. The data set contains two sets of ranking data. In the first set, the subjects are asked to provide a total ranking over 10 items (popular types of sushi). In this set, all subjects are asked to rank the same 10 objects. This set contains 5000 such total rankings.

In the second set of ranking data, a total of 100 types of sushi are considered. We first describe how the 100 types are chosen. The authors in \cite{kamishima2003nantonac} surveyed menu data from 25 sushi restaurants found on the internet. For each type of sushi sold at the restaurant, they counted the number of restaurants that offered the item. From these counts, they derived the probabilities that each item would be supplied. By eliminating unfamiliar or low frequency items, they came up with a list of 100 items. Each subject in this set is asked to rank a subset of 10 items randomly selected from the 100 items, according to the probability distribution described above. This set contains responses from 5000 subjects. 

\noindent In addition, this data set contains demographic information about all the subjects, including their
\begin{enumerate}[label=(\alph*)]
    \item Gender \{Male, Female\} 
    \item Age \{Above 30,  Below 30\} 
    \item Current region of residence  \{East, West\} 
    \item Primary region of residence until 15 years old \{East, West\}. 
\end{enumerate}

\begin{figure}[t!]
  \centering
 \begin{subfigure}[t]{.48\textwidth}
  \centering  
  \includegraphics[height=5cm]{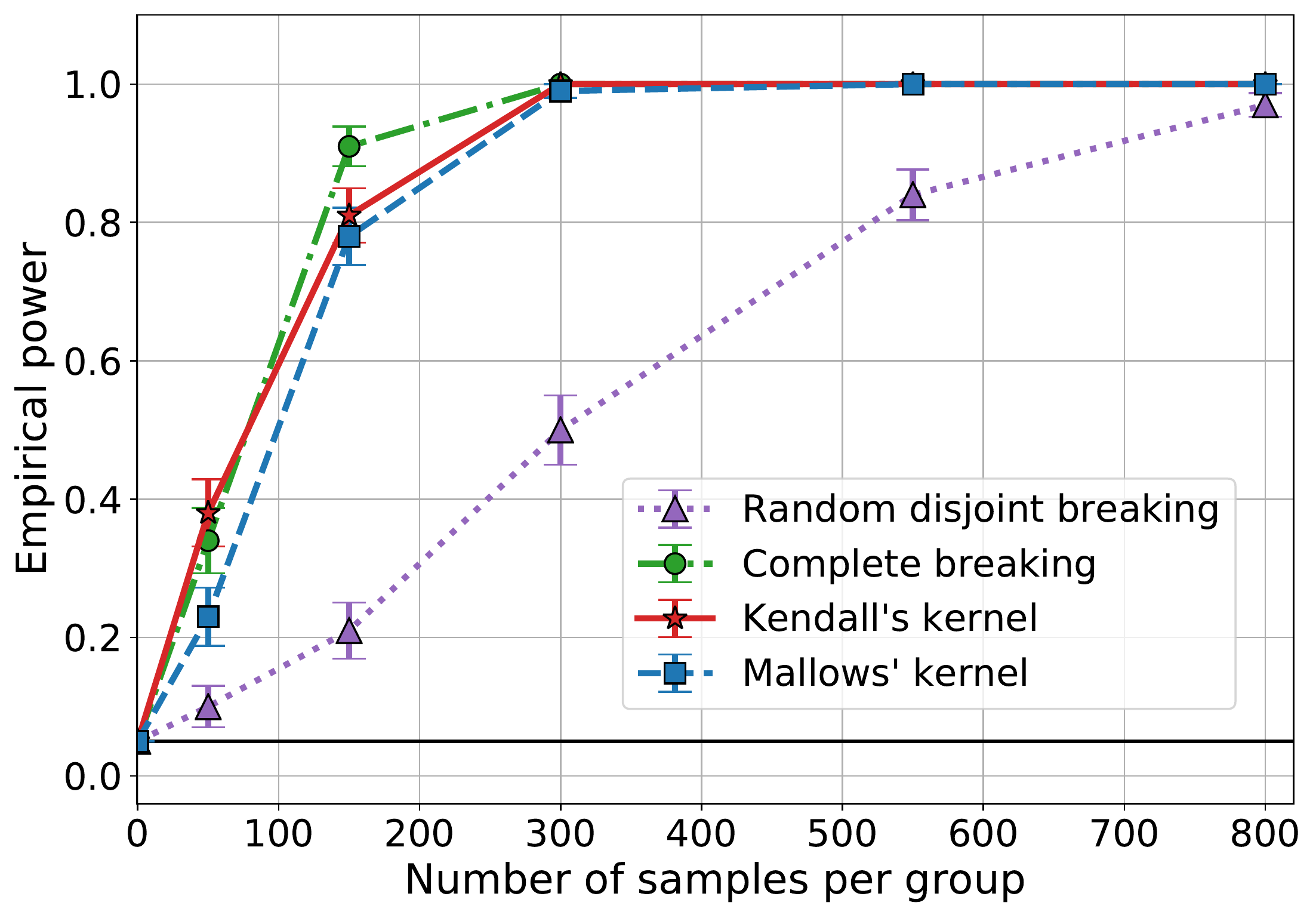}
  \caption{Grouping by gender}
\end{subfigure}%
\begin{subfigure}[t]{.48\textwidth}
  \centering
  \includegraphics[height=5cm]{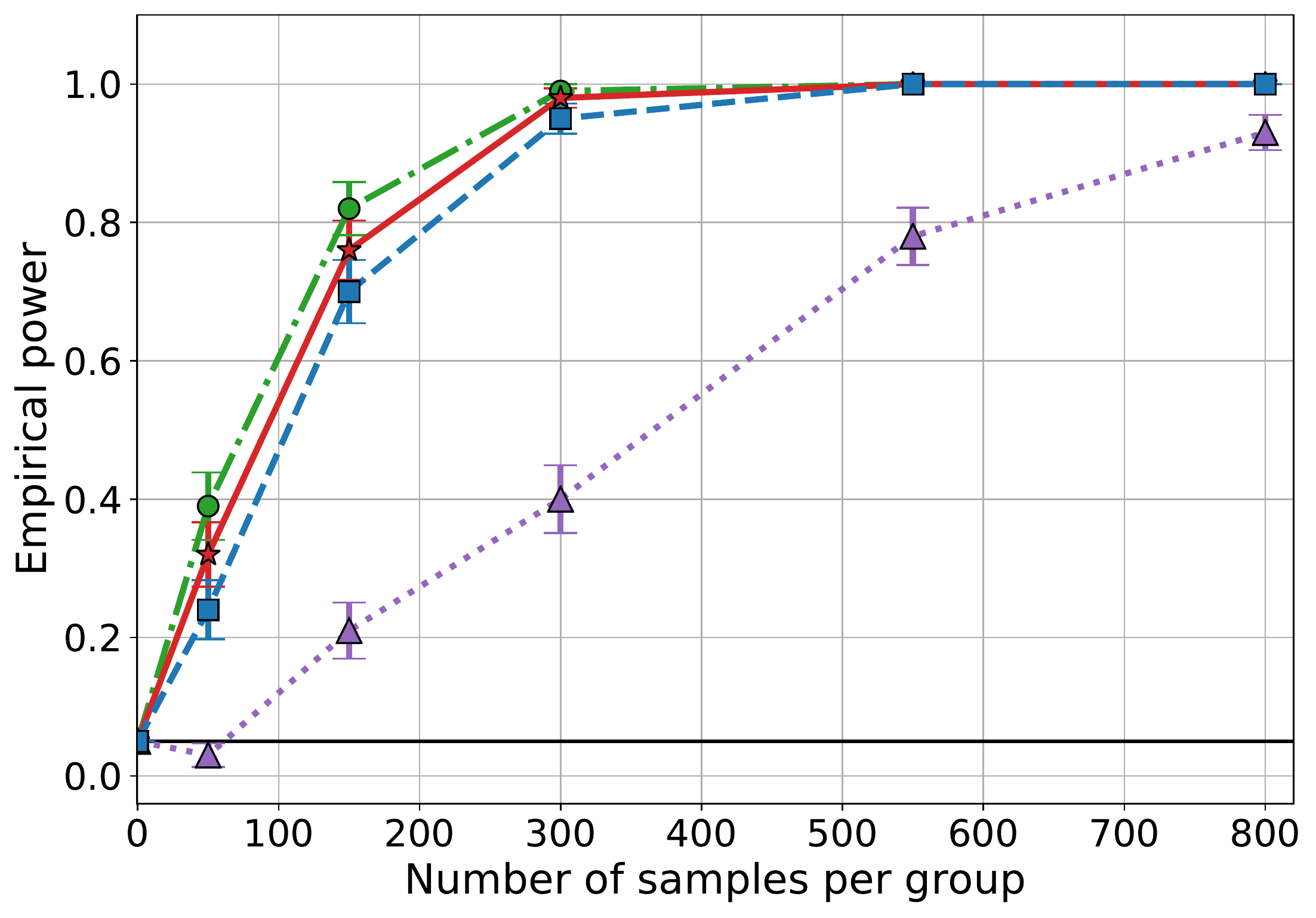}
  \caption{Grouping by age}
\end{subfigure}
\begin{subfigure}[t]{.48\textwidth}
  \centering  
  \includegraphics[height=5cm]{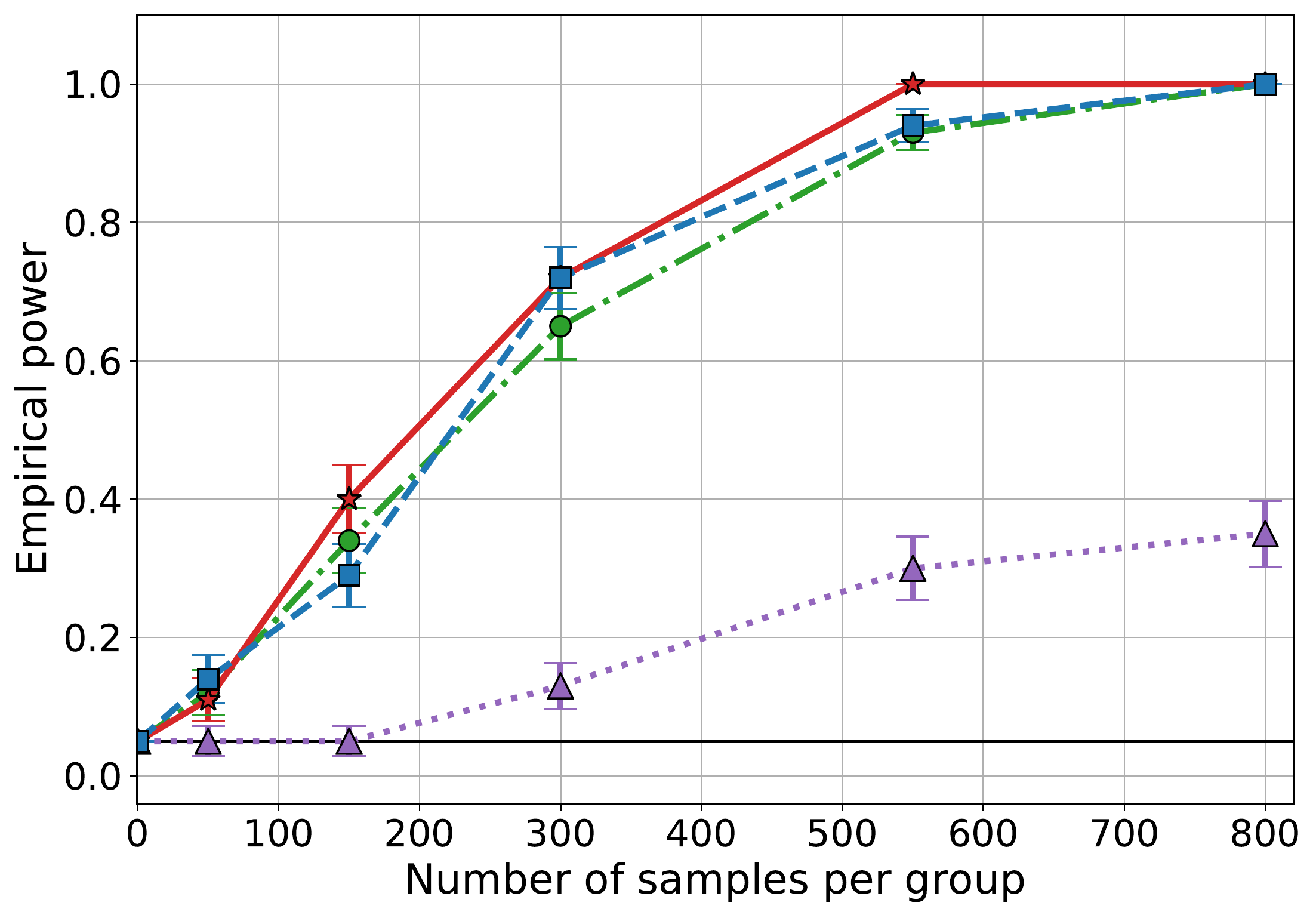}
  \caption{Grouping by primary region of residence until 15 years old }
\end{subfigure}%
\begin{subfigure}[t]{.48\textwidth}
  \centering
  \includegraphics[height=5cm]{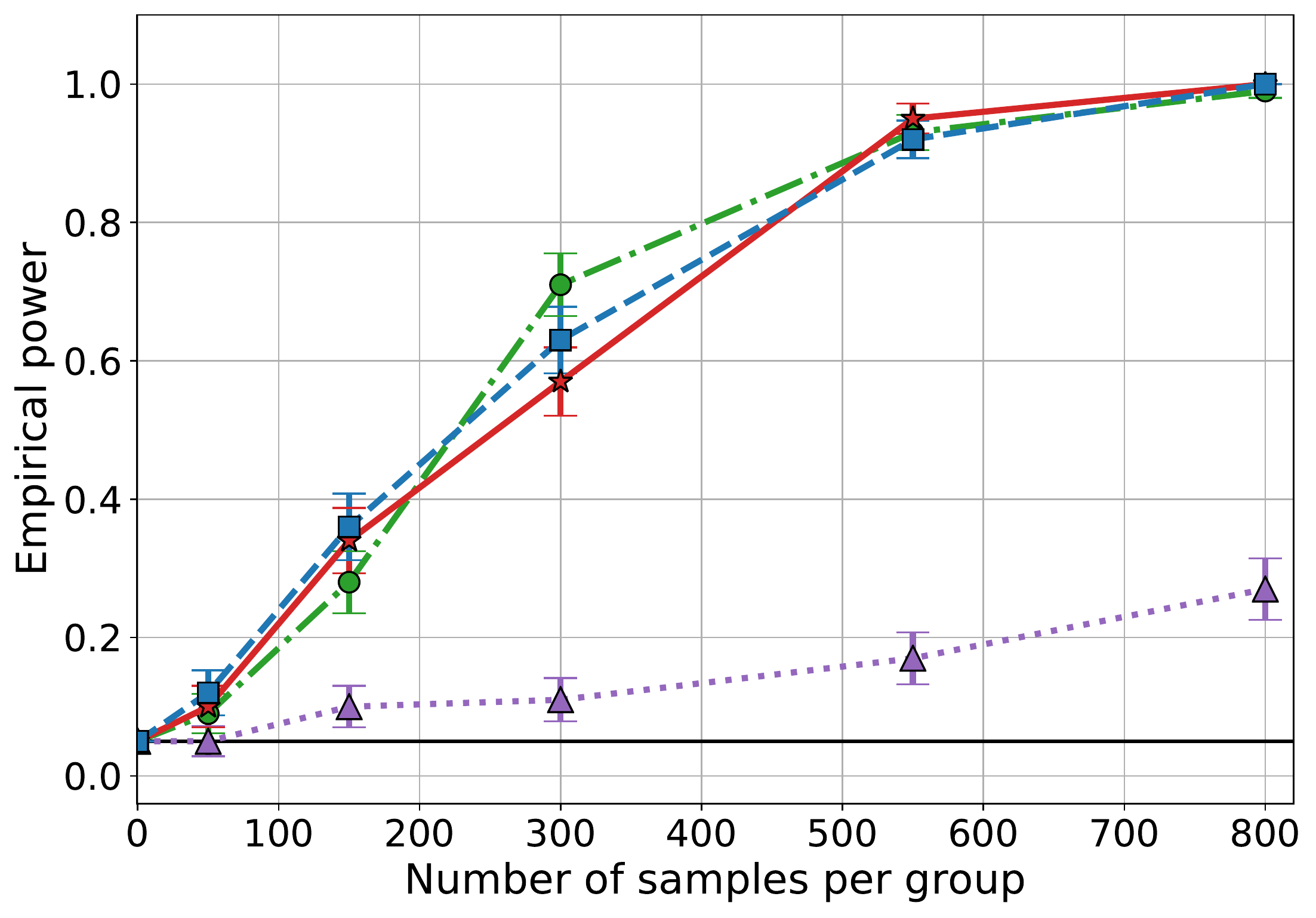}
  \caption{Grouping by current region of residence}
\end{subfigure}
    \caption{\small Empirical power of our test statistic $\stat$ with the permutation testing method described in Algorithm~\ref{permalgo} in testing for difference in sushi preference from the first set of responses with $\items = 10$. The responses obtained comprise of total rankings from each subject. Test results are shown for differences between demographic division based on the information available. Two different rank breaking methods are used for our algorithm, namely, ``Random Disjoint'' and ``Complete''. We also show the results for kernel-based two-sample testing with Kendall's kernel and Mallows' kernel as in  \cite{mania2018kernel}. The x-axis shows the number of samples (total rankings) from each group used to conduct the test and the y-axis shows the empirical power of our test. The test is conducted at a significance level of $0.05$ (indicated by the horizontal line at $y =0.05$). Empirical power is computed as an average over 100 trials. }
  \label{fig:sushi_rank}
\end{figure}

Using our testing algorithm, we test for a difference in preferences across the two sections within each demographic mentioned above. In the first set of experiments, we implement the permutation testing method with our test statistic \eqref{eq:teststat} on the first set of ranking data with $\items = 10$ for each demographic division. We show the results in Figure~\ref{fig:sushi_rank} for two rank-breaking methods, namely, ``Random Disjoint'' and ``Complete''. In addition, we show the results of two kernel-based two-sample testing methods for total rankings designed in \cite{mania2018kernel}, namely, Kendall's kernel and Mallows' kernel.  We randomly sub-sampled $n$ samples from each sub-group of subjects and used $200$ permutations to determine the rejection threshold for the permutation test. In these experiments, we show the empirical power of our testing method, which is the fraction of times our test rejected the null in a total of $100$ trials. We show all the results of using this method on the sushi data set in Figure~\ref{fig:sushi_rank}. \emph{Across each demographic division, our test detects a statistically significant difference in distribution over sushi preferences for the 10 types of sushi included}. Moreover, our testing algorithm with ``Complete'' rank breaking method performs competitively with the kernel-based two-sample testing methods introduced in \cite{mania2018kernel}.

\begin{figure}[t!]
\centering
\begin{subfigure}[t]{.48\textwidth}
  \centering  
  \includegraphics[height=5cm]{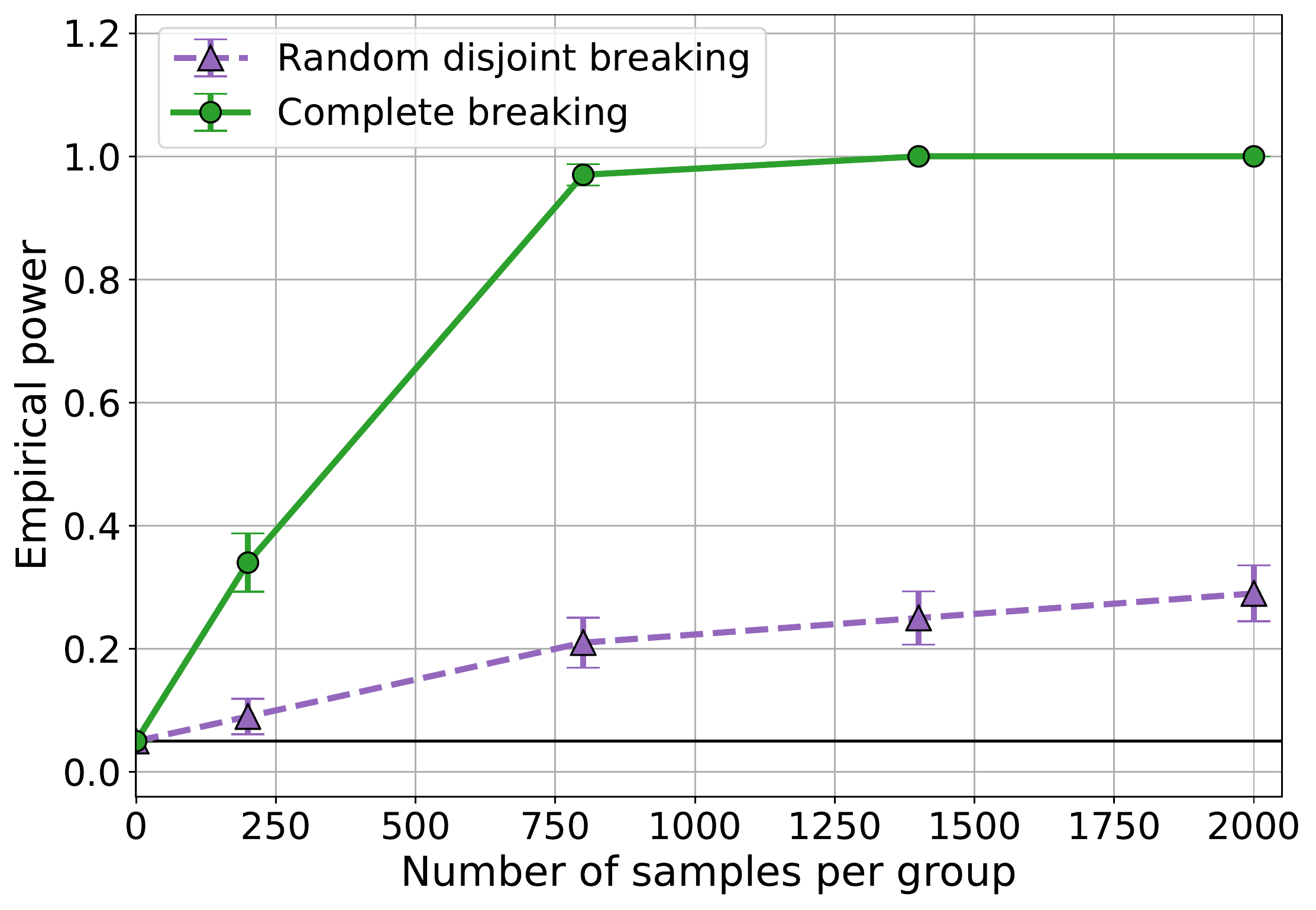}
  \caption{Grouping by gender }
\end{subfigure}%
\begin{subfigure}[t]{.48\textwidth}
  \centering
  \includegraphics[height=5cm]{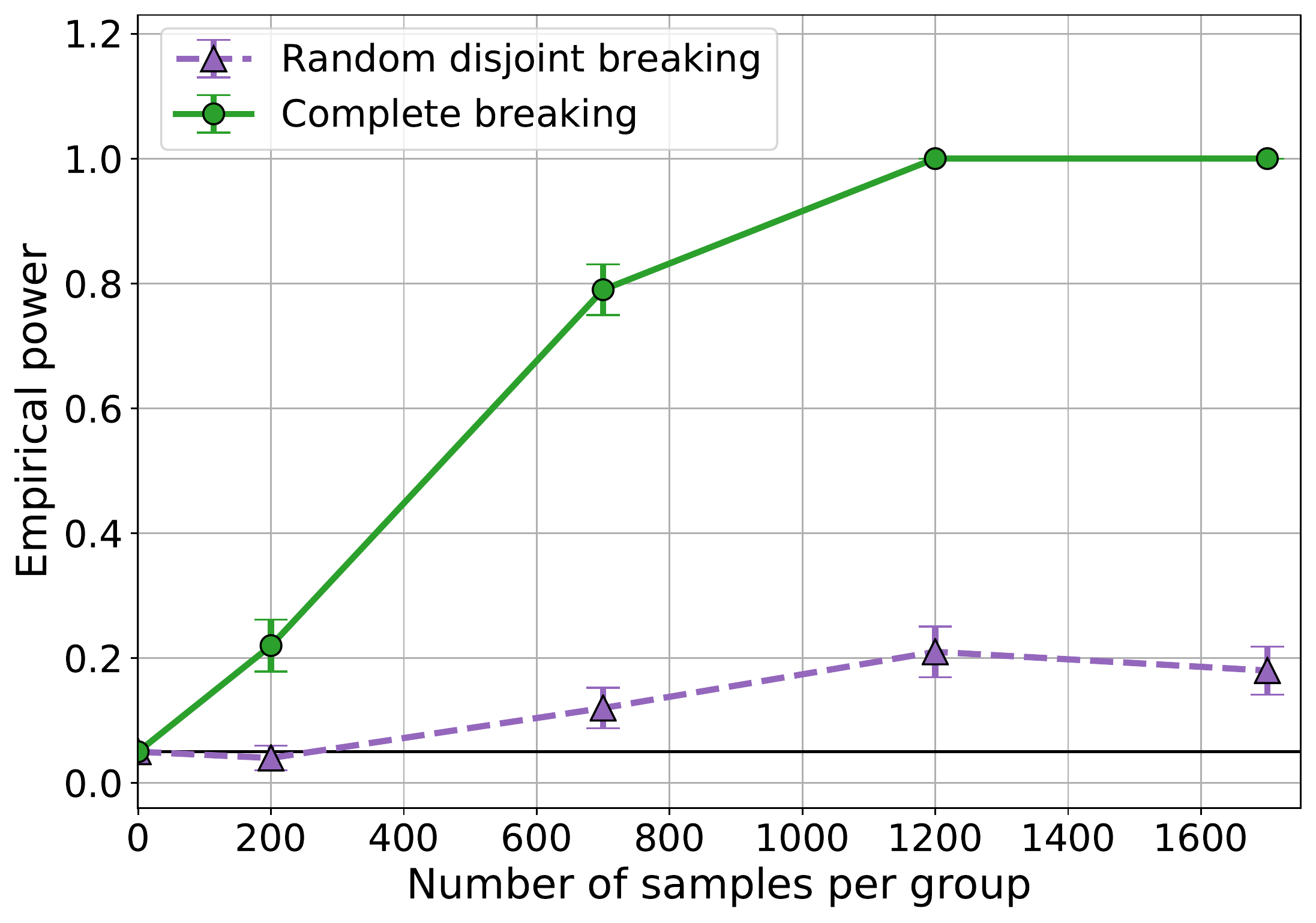}
  \caption{Grouping by age}
\end{subfigure}
\begin{subfigure}[t]{.48\textwidth}
  \centering  
  \includegraphics[height=5cm]{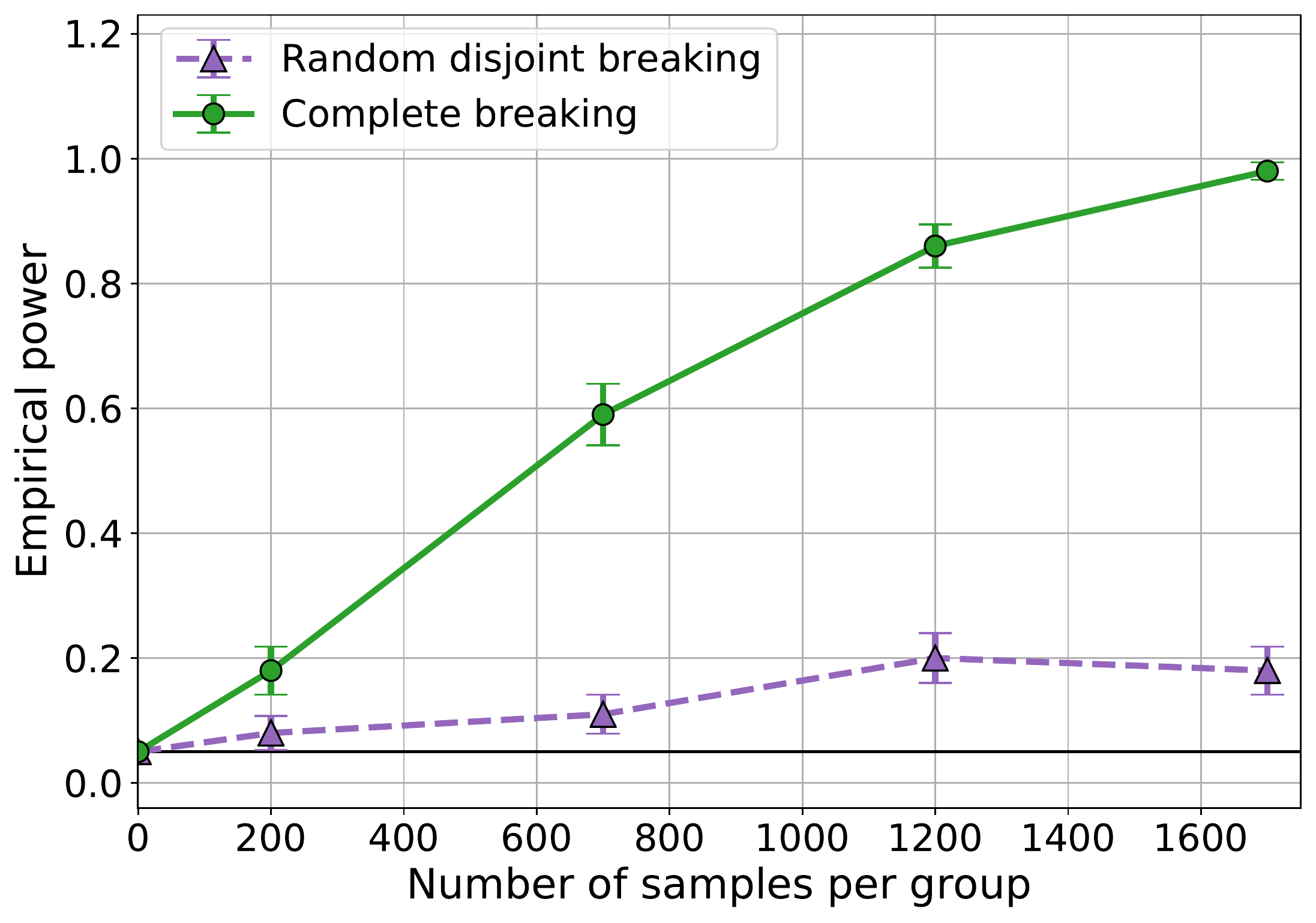}
  \caption{Grouping by primary region of residence until 15 years old }
\end{subfigure}%
\begin{subfigure}[t]{.48\textwidth}
  \centering
  \includegraphics[height=5cm]{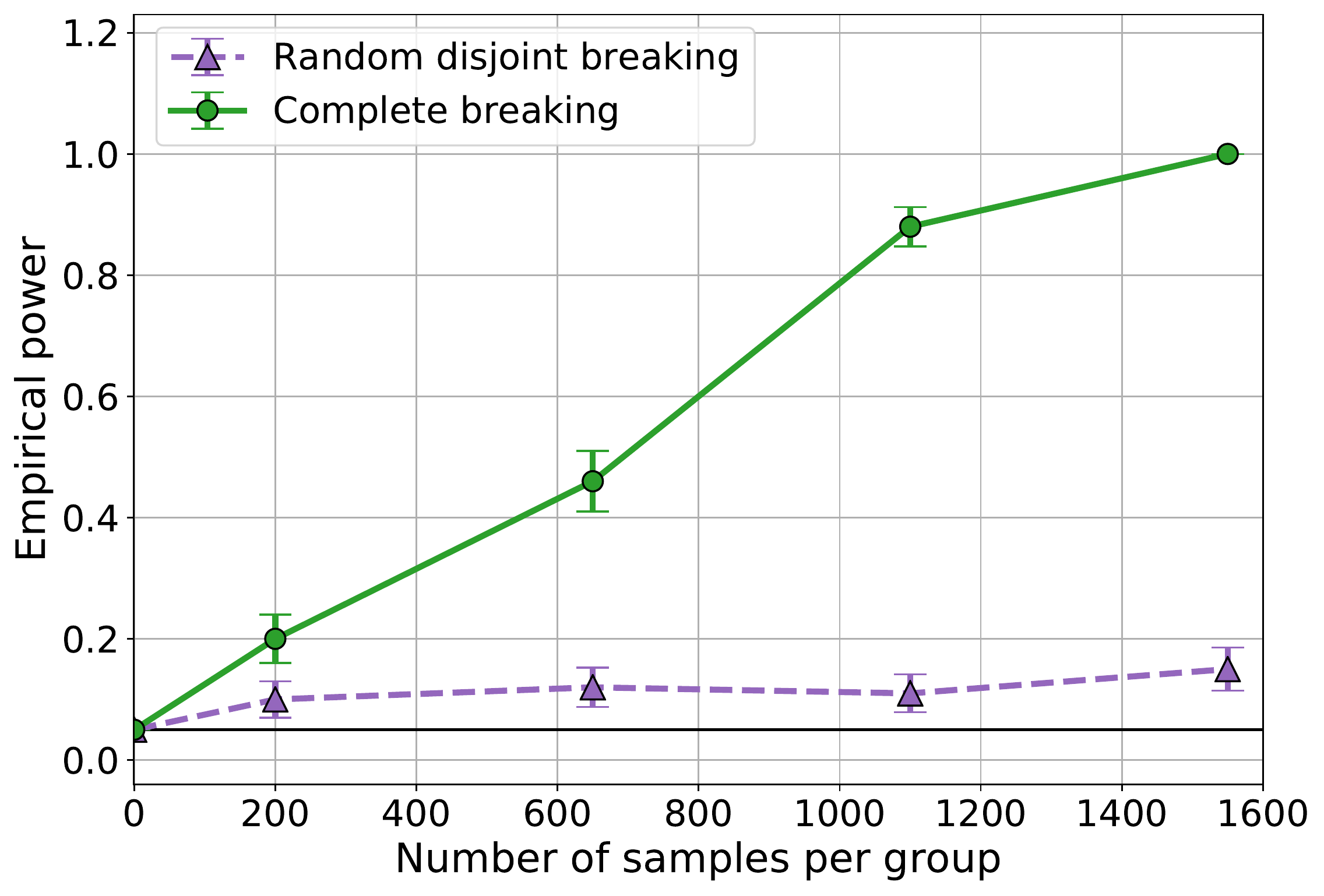}
  \caption{Grouping by current region of residence}
\end{subfigure}
\caption{\small Empirical power of our test statistic $\stat$ with the permutation testing method described in Algorithm~\ref{permalgo} in testing for difference in sushi preference from the first set of responses with $\items = 100$. Test results are shown for differences between demographic division based on the information available. Two different rank breaking methods are used, namely, ``Random Disjoint'' and ``Complete''. The x-axis shows the number of samples (total rankings) from each sub-group used to conduct the test and the y-axis shows the empirical power of our test. The test is conducted at a significance level of $0.05$ (indicated by the horizontal line at $y =0.05$). Empirical power is computed as an average over 100 iterations.
\label{fig:power_d100}}
\end{figure}

We note that since our testing algorithms also work with partial ranking data, our testing algorithms are much more general than the testing algorithms in~\cite{mania2018kernel}, as we demonstrate in our next set of experiments on the second sushi preference data set. We use Algorithm~\ref{permalgo} to test if the preferences of the subjects in the second sushi data set also varies across the different demographics. Recall that this data set has $\items = 100$ items in total but each ranking only ranks a subset of $10$ items. The other details of the experiment are the same as the previous experiment. The results are shown in Figure~\ref{fig:power_d100}. Again, across each demographic, our test detects a statistically significant difference in distribution over sushi preferences for the 100 types of sushi included.

\section{Proofs}
\label{sec:proofs}
This section is devoted to the proofs of our main results. In Section~\ref{sec:proofremupperbound} and Section~\ref{sec:proofupperbound}, we prove the positive results from Section~\ref{sec:upperbound}, and in Section~\ref{sec:proofconverse} we prove the converse results from Section~\ref{sec:lowerbound}. Lastly, Sections~\ref{sec:proofalgo1}-\ref{sec:proofalgo2} are devoted to proofs of results under the partial (or total) ranking setting mentioned in Section~\ref{sec:results_partial}.

Throughout these and other proofs, we use the notation ${\const, \const', \const_0, \const_1}$ and so on to denote positive constants whose values may change from line to line.

\subsection{Proof of Corollary~\ref{rem:upperbound}}
\label{sec:proofremupperbound}

In this section we present the complete proof of Corollary~\ref{rem:upperbound}. We first present the proof for the random-design setup described in Corollary~\ref{rem:upperbound} and then specialise the proof in Section~\ref{sec:proofupperbound} to the per-pair fixed-design setup in Theorem~\ref{thm:upperbound}. To prove our result, we analyse the expected value and the variance of the test statistic $\stat $ in Algorithm~\ref{testalgo} in the following two lemmas. Recall that under the random-design setup $\kpij, \kqij$ are distributed independently and identically according to some distribution $\kdist$ that satisfies the conditions in \eqref{eq:momentprop}. 
\begin{lemma}
    For $\stat$ as defined in Algorithm~\ref{testalgo}, with $\kpij, \kqij \stackrel{\textnormal{iid}}{\sim} \kdist$, under the null $\E_{\nullh}[\stat] = 0$ and under the alternate,  
    \begin{align*}
        \E_{\alt}[\stat] \geq  \const\mean \frobnorm{\popp-\popq}^2.
    \end{align*}
    \label{lem:expectation}
\end{lemma}
\noindent The proof of Lemma~\ref{lem:expectation} is provided in Section~\ref{sec:prooflemmaexp}. Now, with a view to applying Chebyshev's concentration inequality, we bound the variance of $\stat$. 
\begin{lemma}
      For $\stat$ as defined in Algorithm~\ref{testalgo}, with $\kpij, \kqij \stackrel{\textnormal{iid}}{\sim} \kdist$, where $\kdist$ obeys the conditions described in \eqref{eq:momentprop}, under the null 
      \begin{align*}
          \Var_{\nullh}{[\stat]} \leq 24\items^2,
      \end{align*} 
      and under the alternate, 
      \begin{align*}
      \Var_{\alt}{[\stat]} \leq 24\items^2 + 8\mean\frobnorm{\popp-\popq}^2 + \const'\mean^2\frobnorm{\popp-\popq}^2
\end{align*}
where $\const'>0$ is a constant. 
      \label{lem:variance}
\end{lemma}
\noindent The proof for Lemma~\ref{lem:variance} is provided in Section~\ref{sec:prooflemmavar}. We now have to control Type I error and Type II error. Using one-sided Chebyshev's inequality for the test statistic $\stat$, which has $\Ebb_{\nullh}[\stat] = 0$, we derive an upper bound on Type I error as follows
\begin{align}
    \Pbb_{\nullh}(\stat \geq \threshold) \leq \dfrac{\Var_{\nullh}[\stat]}{\Var_{\nullh}[\stat]+\threshold^2}.
    \label{eq:null}
\end{align}
Observe that if $\threshold  = 11\items$ then the Type I error is upper bounded by $\frac{1}{6}$. 
In addition, if Type I error is required to be at most $\constfactor$, then we set the threshold equal to $\items \sqrt{\dfrac{24(1-\constfactor)}{\constfactor}}$. We now move to controlling the Type II error of the testing algorithm. We again invoke Chebyshev's inequality as follows 
\begin{align}
    \P_{\alt}(\stat < \threshold) \leq \dfrac{\Var_{\alt}[\stat]}{\Var_{\alt}[\stat]+ (\E_{\alt}[\stat]-\threshold)^2}.
     \label{eq:alternative}
\end{align}
\noindent To guarantee that Type II error is at most $ \frac{1}{6}$, we substitute the bounds on $\E_{\alt}[\stat], \Var_{\nullh}[\stat], \Var_{\alt}[\stat]$ from Lemma~\ref{lem:expectation} and Lemma~\ref{lem:variance} in \eqref{eq:alternative} to get the sufficient  condition
\begin{align*}
    \nonumber 5(24\items^2 + 8\mean\frobnorm{\popp-\popq}^2 + \const'\mean^2\frobnorm{\popp-\popq}^2) &\leq (\const\mean\frobnorm{\popp-\popq}^2 -11\items)^2 \\
    \nonumber 40\mean\frobnorm{\popp-\popq}^2 + 22c\items\mean\frobnorm{\popp-\popq}^2 + 5\const'\mean^2\frobnorm{\popp-\popq}^2 &\leq \const^2\mean^2\frobnorm{\popp-\popq}^4 + \items^2.
    \end{align*}
This condition yields
\begin{align}
     40 + 22\const \items + 5\const'\mean &\leq \const^2\mean \frobnorm{\popp-\popq}^2.
     \label{eq:powercond}
\end{align}
Recall that under the alternate $\frac{1}{\items}\frobnorm{\popp-\popq} \geq \distance$. 
According to the final condition derived here \eqref{eq:powercond}, under the regime $\mean > \items$, we have control over total probability of error if $\distance^2\items^2 \geq \const'$ for some constant $\const' > 0$. Under the regime $\mean \leq \items $, the condition \eqref{eq:powercond} simplifies as 
\begin{align}
     \distance^2 \geq \dfrac{\const''}{\mean\items},
    \label{eq:upper}
\end{align}
where $\const''>0$ is some constant.
This gives the sufficient condition to control total probability of error (sum of Type I error and Type II error) to be at most $\frac{1}{3}$ under the setting where $\kpij, \kqij \stackrel{\textnormal{iid}}{\sim} \kdist$. 

\subsubsection{Proof of Lemma~\ref{lem:expectation}}
\label{sec:prooflemmaexp}
We now prove the bounds on the expected value of the test statistic defined in Algorithm~\ref{testalgo}. Recall that for each $(i,j)$, given $\kpij, \kqij$, we have $\matx_{ij} \sim \textnormal{Bin}(\kpij,\pij{})$ and $\maty_{ij} \sim \textnormal{Bin}(\kqij,\qij{}) $. Also, $\kpij,\kqij \stackrel{\textnormal{iid}}{\sim} \kdist$ wherein $\E[\kpij]=\mean, \Var[{\kpij}]=\stddev^2, \pr(\kpij=1)=p_1$ and $\kdist$ obeys \eqref{eq:momentprop}. We denote the vector of $\kpij$ and $\kqij$ for all $(i,j)$ by $\textbf{\samples}^\distra$ and $\textbf{\samples}^\distrb$  respectively. Now, the conditional expectation of $\stat$ is expressed as
\begin{align}
    \mathbb{E}\,[\,\stat\,|\, \textbf{\samples}^\distra,\textbf{\samples}^\distrb\,] = \sum_{i=1}^\items\sum_{j=1}^\items \dfrac{\mathbb{I}_{ij}\kpij\kqij}{\kpij+\kqij}(\pij{}-\qij{})^2.
    \label{eq:exp}
\end{align}

Using the law of total expectation, we have 
\begin{align*}
     \E\,[\,\stat\,] &= \E\,[\,\E\,[\,\stat\,|\,\textbf{\samples}^\distra,\textbf{\samples}^\distrb\,]\,]\\
      & = \sum_{i=1}^\items\sum_{j=1}^\items \E\left[\dfrac{\mathbb{I}_{ij}\kpij\kqij}{\kpij+\kqij}(\pij{}-\qij{})^2\right]\\
      & = \E\left[\dfrac{\mathbb{I}_{ij}\kpij\kqij}{\kpij+\kqij}\right]\frobnorm{\popp-\popq}^2
\end{align*}
Clearly, $\E_{\nullh}[\stat] = 0$. To find a lower bound for $\E_{\alt}[\stat]$, we first note that 
\begin{align}
\nonumber     \E\left[\dfrac{\mathbb{I}_{ij}\kpij\kqij}{\kpij+\kqij}\right] &= \E\left[\dfrac{\mathbb{I}_{ij}^{0}\kpij\kqij}{\kpij+\kqij}\right] - 2\sum_{\samples\in [\items]} \Pbb(\kpij=1,\kqij=\samples)\dfrac{\samples}{\samples + 1} + \dfrac{1}{2}\Pbb(\kpij=1,\kqij=1)\\
     & \geq \E\left[\dfrac{\mathbb{I}_{ij}^{0}\kpij\kqij}{\kpij+\kqij}\right] -2\pone.
     \label{eq:indicator1}
\end{align}
where $\mathbb{I}_{ij}^0 = \mathbb{I}(\kpij>0)\times \mathbb{I}(\kqij>0)$. Furthermore, we see that for any event $E$, 
\begin{align}
     \E\left[\dfrac{\mathbb{I}_{ij}^0\kpij\kqij}{\kpij+\kqij}\right] \geq  \E\left[\dfrac{\mathbb{I}_{ij}^0\kpij\kqij}{\kpij+\kqij}\;|\,E\,\right]\pr(E).
     \label{eq:indicator2}
\end{align}
We define the event $E$ as 
\begin{equation}
\begin{aligned}
    \mean - \const\stddev &\leq \kpij \leq \mean + \const\stddev, \textnormal{and}\\
    \mean - \const\stddev &\leq \kqij \leq \mean + \const\stddev
    \label{eq:indicator3}
\end{aligned}
\end{equation}
with some constant $\const >1$ such that $\mean - \const\stddev >0$. 
Using Chebyshev's inequality, we get that $\pr(E) \geq (1-\frac{1}{\const^2})^2$. Finally, we combine \eqref{eq:indicator1}, \eqref{eq:indicator2} and \eqref{eq:indicator3}, to get 
\begin{align}
     \E\left[\dfrac{\mathbb{I}_{ij}\kpij\kqij}{\kpij+\kqij}\right] \geq \dfrac{(\mean - \const\stddev)^2}{2(\mean + \const\stddev)}\left(1-\dfrac{1}{\const^2}\right)^2 - 2\pone.
     \label{eq:explowbound}
\end{align}
Since $\kdist$ obeys the conditions in \eqref{eq:momentprop}, we have $\mean \geq \const_1p_1$ and $\mean \geq \const_2\stddev$. Therefore, there is a constant $\const>0$ that depends on $\const_1, \const_2$, such that $\E[\stat] \geq \const\mean \frobnorm{\popp-\popq}^2$.
This proves Lemma~\ref{lem:expectation}.

\subsubsection{Proof of Lemma~\ref{lem:variance}}
\label{sec:prooflemmavar}
To analyse the variance of the test statistic $\stat$, we note that pairwise-comparisons for each pair are obtained independently. This allows us to compute the variance for each pair $(i,j)$ separately, as variance of sum is equal to the sum of variances. The following analysis of the variance of the test statistic $\stat$ applies under both the null and the alternate. The law of total variance states that
\begin{align}
    \Var[\stat] = \E[\,\Var[\stat|\textbf{\samples}^\distra, \textbf{\samples}^\distrb]\,] + \Var[\,\E[\stat|\textbf{\samples}^\distra, \textbf{\samples}^\distrb]\,]. 
    \label{eq:lawtotalvar}
\end{align}
We evaluated the term $\Var{[\stat|\textbf{\samples}^\distra, \textbf{\samples}^\distrb]}$, present in the expression above, in Wolfram Mathematica. We show the output here, 
\begin{align}
     \nonumber \Var[\stat|\textbf{\samples}^\distra, \textbf{\samples}^\distrb] & \le \sum_{i=1}^\items\sum_{j=1}^\items \dfrac{2\mathbb{I}_{ij}\kpij(\kpij-1)\kqij(\kqij-1)}{(\kpij-1)^2(\kqij-1)^2(\kpij+\kqij)^2}\Big(\kqij(\kqij-1)\pij{4}(3-2\kpij)\\&+  2\pij{3}\kqij(\kqij-1)(-2+2\qij{}\kpij -  2\qij{}+\kpij) \nonumber  \\&+ \nonumber 2\pij{}\qij{}(\kpij-1)(\kqij-1)(1+2\qij{2}\kpij-\qij{}-2\qij{}\kpij+\qij{}\kqij) 
     \\&+ \nonumber \pij{2}(\kqij-1)(2\qij{}(\kpij-1)(\kpij-1-2\kqij) + \kpij- 2\qij{2}(\kpij-1)(\kpij+\kqij-1)) \\\nonumber &-  \qij{2}(\qij{}-1)\kpij(\kpij-1)(1-3\qij{}+2\qij{}\kqij)\Big) \\ \nonumber  &\le \sum_{i=1}^\items\sum_{j=1}^\items \dfrac{8\mathbb{I}_{ij}}{(\kpij+\kqij)^2}\Big(\kpij(\kpij-1)(\kqij-1)(2\pij{}(\pij{}-\qij{})^2) \\&+ \nonumber  \kqij(\kqij-1)(\kpij-1)(2\qij{}(\pij{}-\qij{})^2) \\&+ \nonumber 2\pij{}\qij{}(\kpij-1)(\kqij-1)(1-\pij{})(1-\qij{}) \\&+ \pij{2}\kqij(\kqij-1)(1-\pij{})^2 + \qij{2}\kpij(\kpij-1)(1-\qij{})^2 \Big).
     \label{eq:var1}
\end{align}

\noindent Applying the trivial upper bound $\pij{}\leq 1 , \qij{}\leq 1\; \forall \; (i,j)$, we get
\begin{align}
      \Var[\stat|\textbf{\samples}^\distra, \textbf{\samples}^\distrb] &\le \sum_{i=1}^\items\sum_{j=1}^\items 8\mathbb{I}_{ij}\left(\dfrac{ \kpij\kqij}{(\kpij+\kqij)}(\pij{}-\qij{})^2 + 3\right)
      \label{eq:var2}
\end{align}
Following this, we evaluate the first term on the right hand side of \eqref{eq:lawtotalvar} as
\begin{align}
    \E[\,\Var[\stat|\textbf{\samples}^\distra, \textbf{\samples}^\distrb]\,] \le 24\items^2 + 8\frobnorm{\popp-\popq}^2 \E\left[\dfrac{\mathbb{I}_{ij}\kpij\kqij}{\kpij+\kqij}\right].
     \label{eq:var3}
\end{align}
To further simplify the upper bound in \eqref{eq:var3}, we observe that 
\begin{align}
    \E\left[\dfrac{\mathbb{I}_{ij}\kpij\kqij}{\kpij+\kqij}\right] \leq \half\E[\max\{\kpij,\kqij\}]. 
    \label{eq:var4}
\end{align}
We exploit the independence of $\kpij, \kqij$ to get the CDF of $\max\{
\kpij, \kqij\}$ as $$\Pbb(\max\{\kpij, \kqij\} \leq x) = \Pbb(\kpij\leq x)\Pbb(\kqij \leq x).$$
Through the CDF, we derive the PDF as 
\begin{align}
    \nonumber\Pbb(\max\{\kpij, \kqij\} = x) &= \Pbb(\max\{\kpij, \kqij\} \leq x) - \Pbb(\max\{\kpij, \kqij\} \leq x-1)\\
   \nonumber &=  \Pbb(\kpij\leq x)^2 - \Pbb(\kpij \leq x-1)^2 \\
   \nonumber &= \Pbb(\kpij=x)(\Pbb(\kpij\leq x) + \Pbb(\kpij\leq x-1))\\
    &\leq 2\Pbb(\kpij=x)
    \label{eq:pdfmax}
\end{align}
We substitute this inequality in \eqref{eq:var4} to get 
\begin{align}
    \E\left[\dfrac{\mathbb{I}_{ij}\kpij\kqij}{\kpij+\kqij}\right] \leq \mean. 
\end{align}
As a result, following from \eqref{eq:var3}, we have 
\begin{align}
    \E[\,\Var[\stat|\textbf{\samples}^\distra, \textbf{\samples}^\distrb]\,] \le 24\items^2 + 8\mean\frobnorm{\popp-\popq}^2.
    \label{eq:expofvar}
\end{align}
Now, the remaining (second) term on the right hand side of \eqref{eq:lawtotalvar} is
\begin{align}
    \nonumber \Var[\,\E[\stat|\textbf{\samples}^\distra, \textbf{\samples}^\distrb]\,] &= \sum_{i=1}^\items\sum_{j=1}^\items \Var\left[\dfrac{\mathbb{I}_{ij}\kpij\kqij}{(\kpij+\kqij)}\right] (\pij{}-\qij{})^4\\
    &\leq  \Var\left[\dfrac{\mathbb{I}_{ij}\kpij\kqij}{(\kpij+\kqij)}\right] \sum_{i=1}^\items\sum_{j=1}^\items (\pij{}-\qij{})^2
     \\&  \leq  \Var\left[\dfrac{\mathbb{I}_{ij}\kpij\kqij}{(\kpij+\kqij)}\right]\frobnorm{\popp-\popq}^2.
     \label{eq:var5}
\end{align}
To bound the variance term in the previous equation, we see that 
\begin{align}
     \Var\left[\dfrac{\mathbb{I}_{ij}\kpij\kqij}{(\kpij+\kqij)}\right]& \nonumber =  \E\left[\left(\dfrac{\mathbb{I}_{ij}\kpij\kqij}{(\kpij+\kqij)}\right)^2\right] - \E\left(\left[\dfrac{\mathbb{I}_{ij}\kpij\kqij}{(\kpij+\kqij)}\right]\right)^2\\
     &\nonumber\stackrel{(a)}{\leq} \dfrac{1}{4} \E\left[\left(\max\{\kpij,\kqij\}\right)^2\right] - \const\mean^2 \\
     &\nonumber\stackrel{(b)}{\leq} \half\E[(\kpij)^2] - \const\mean^2\\
     &\nonumber = \half(\mean^2 + \stddev^2)-\const\mean^2\\
     &\stackrel{(c)}{\leq}\left(\half +\frac{1}{2\const_2^2}-\const\right)\mean^2 = \const'\mean^2,
     \label{eq:var6}
\end{align}
where inequality (a) follows from \eqref{eq:explowbound}, inequality (b) follows similarly to the result in \eqref{eq:pdfmax}, and inequality (c) is a result of \eqref{eq:momentprop}. Thus, the upper bound in \eqref{eq:var5} becomes 
\begin{align}
    \Var[\,\E[\stat|\textbf{\samples}^\distra, \textbf{\samples}^\distrb]\,] \leq \const'\mean^2\frobnorm{\popp-\popq}^2.
    \label{eq:varofexp}
\end{align}
Finally, we put together the terms in \eqref{eq:lawtotalvar} by combining \eqref{eq:expofvar} and \eqref{eq:varofexp} to get the desired upper bound on the variance of the test statistic under the alternate hypothesis, which is 
\begin{align}
     \Var{[\stat]} \leq 24\items^2 + 8\mean\frobnorm{\popp-\popq}^2 + \const'\mean^2\frobnorm{\popp-\popq}^2.
     \label{eq:final_var}
\end{align}

Additionally, to obtain the upper bound on the variance of the test statistic under the null, we substitute $\frobnorm{\popp-\popq}$ as zero in \eqref{eq:final_var}.  This completes the proof of Lemma~\ref{lem:variance}.

\subsection{Proof of Theorem~\ref{thm:upperbound}}
\label{sec:proofupperbound}
In this proof, we first specialise the statements of Lemma~\ref{lem:expectation} and Lemma~\ref{lem:variance} to the per-pair fixed-design setup where $\kpij = \kqij =\samples \; \forall \; (i,j)\in [\items]$, for some positive integer $\samples>1$. Under this setting, following from \eqref{eq:exp}, we have 
\begin{align}
    \Ebb[\stat] =  \sum_{i=1}^\items\sum_{j=1}^\items \half\mathbb{I}(\samples >1)\samples(\pij{}-\qij{})^2 = \half\samples\frobnorm{\popp -\popq}^2.
    \label{eq:expfixed}
\end{align}
Similarly, we note that in \eqref{eq:lawtotalvar} we have that $\Var[\,\E[\stat|\textbf{\samples}^\distra, \textbf{\samples}^\distrb]\,] = 0$, which in combination with \eqref{eq:var3} implies that 
\begin{align}
    \Var{[\stat]} \leq 24\items^2 + 4\samples\frobnorm{\popp-\popq}^2.
    \label{eq:varfixed}
\end{align}
Now, invoking Chebyshev's inequality as described in \eqref{eq:null} and \eqref{eq:alternative} to control Type I and Type II error at level $\frac{1}{6}$, we set the threshold as $11\items$ to get the sufficient condition as 
\begin{align}
    \distance^2 \geq \frac{\const}{\samples\items}
\end{align}
for some positive constant $\const$. This proves Theorem~\ref{thm:upperbound}.

\subsection{Proof of converse results}
\label{sec:proofconverse}
In this section we prove all the claims made in Section~\ref{sec:lowerbound}. We begin with some background.

\subsubsection{Preliminaries for proof of lower bound}
\label{sec:preliminaries}

We begin by briefly introducing the lower bound technique applied in Theorem~\ref{thm:lbmst} and Theorem~\ref{thm:lbbtl}. The main objective of the proof is to construct a set of null and alternate such that the minimax risk of testing defined in \eqref{eq:risk} is lower bounded by a constant. To lower bound the minimax risk, we analyse the $\chisq$ distance between the resulting distributions of the null and the alternate.  We construct the null and alternate as follows. Let $\popp_0 = [\half]^{\items \times \items}$. Under the null, we fix $\popp = \popq = \popp_0$ and under the alternate, $\popp = \popp_0, \popq \in \mset$ where $\mset$ is a set of matrices from the model class $\modelclass$ to be defined subsequently. We assume a uniform probability measure over $\mset$. The set $\mset$ is chosen such that $\frac{1}{\items}\frobnorm{\popp_0 -\popq} = \distance$ for all $\popq\in \mset$.  

In our problem setup, we observe matrices $\matx, \maty$ wherein each element is the outcome of $\samples$ observations of the corresponding Bernoulli random variable. For each pair $(i,j)$, we have $\matx_{ij}\sim\Bin(\samples , \pij{}), \maty_{ij}\sim\Bin(\samples, \qij{})$. For simplicity of notation, we will denote the matrix distribution corresponding to the pairwise-comparison probability matrix $\popp_0$ by $\Pbb_0$, that is, $\matx \sim \Pbb_0$ when $\popp = \popp_0$, and $\maty \sim \Pbb_0$ when $\popq = \popp_0$. For the case where $\maty_{ij}\sim\Bin(\samples, \qij{})$ and $\popq \sim\textnormal{Unif}(\mset)$, we denote the resulting matrix distribution as $\maty \sim \Pbb_{\mset}$. We now have all the parts required to derive the $\chisq$ divergence between the null and the alternate defined in this section. \\

The $\chisq$ divergence between the distribution of $\matx, \maty$ under the null and the distirbution of $\matx, \maty$ under the alternate is given by
\begin{align}
   \nonumber \chisq((\matx,\maty)_{\nullh}, (\matx, \maty)_{\alt}) 
    &= \chisq(\matx_{\nullh}, \matx_{\alt}) + \chisq(\maty_{\nullh}, \maty_{\alt}) +  \chisq(\matx_{\nullh}, \matx_{\alt})\chisq(\maty_{\nullh}, \maty_{\alt})\\     
 \nonumber    & = \chisq(\Pbb_0,\Pbb_0)  + \chisq(\Pbb_0, \mathbb{P}_{\mset}) + \chisq(\Pbb_0, \Pbb_0)\chisq(\Pbb_0, \mathbb{P}_{\mset})\\
    & = \chisq(\Pbb_0, \P_{\mset}).
    \label{eq:goodnessoffit}
\end{align}
This reduces our two-sample testing problem into a goodness of fit testing problem for the given model class, where the null distribution is given by $\Pbb_0$ and the alternate distribution is given by $\Pbb_{\mset}$. This goodness of fit testing problem is written as 
\begin{align}
    \begin{split}
        \nullh &: \popp = \popp_0   \\
        \alt &: \popp \sim \textnormal{Unif}(\mset),
    \end{split}
    \label{eq:goodnessoffit_test}
\end{align}
where $\popp_0 = \left[\half\right]^{\items \times \items}$.


Continuing with the reduction in \eqref{eq:goodnessoffit} and \eqref{eq:goodnessoffit_test}, Le Cam's method for testing states that the minimax risk \eqref{eq:risk} for the hypothesis testing problem in \eqref{eq:goodnessoffit_test}, is lower bounded as  (Lemma~3 in \cite{collier2017minimax})
\begin{align}
    \risk_\modelclass \geq\half\left( 1 - \sqrt{\chisq(\Pbb_0, \P_{\mset})}\right).
    \label{eq:mmrisk}
\end{align}
Therefore, if the $\chisq$ divergence is upper bounded by some constant $\const < 1$, then no algorithm can correctly distinguish between the null and the alternate with probability of error less than $\half(1 - \sqrt{\const})$. Consequently, by deriving the value of $\distance$ corresponding to  $\const = \frac{1}{9}$, we will get the desired lower bound on the critical radius defined in \eqref{eq:criticalrad} for the two-sample testing problem in \eqref{eq:test}.

We now delve into the technical part of the proof in which we derive the $\chisq$ divergence between $\Pbb_0$ and $\Pbb_{\mset}$. 
For a probability distribution $\Pbb_0$ and a mixture probability measure $\Pbb_{\mset}$, we know (from Lemma~7 in \cite{carpentier2018minimax}) that
\begin{align}
    \chisq(\Pbb_0 ,\P_{\mset}) = \E _{(\popq, \popq')\sim\textnormal{Unif}(\mset)}\left(\int \dfrac{ d\Pbb_{\popq}d\Pbb_{\popq'}}{d\Pbb_0} \right) - 1.
\end{align}
Here $\E _{(\popq, \popq')\sim\textnormal{Unif}(\mset)}$ denotes the expectation with respect to the distribution of the pair $(\popq, \popq')$ where $\popq$ and $\popq'$ are sampled independently and uniformly at random from the set $\mset$ (with replacement). According to the null and alternate construction described in the beginning of this section, recall that $\matx \sim \Pbb_0$ implies that $\matx_{ij} \sim \Bin(\samples, \half) \; \forall \; (i,j)$. Similarly $\matx\sim\Pbb_\popq$ implies that $\matx_{ij} \sim \Bin(\samples, \qij{}) \; \forall \; (i,j)$. 
With this information, we simplify the $\chisq$ divergence as 
\begin{align}
 \chisq(\Pbb_0 ,\P_{\mset}) = \E_{(\popq,\popq')\sim \textnormal{Unif}(\mset)} \sum_{v \in V}  \left(\prod_{i=1}^\items\prod_{j=1}^\items \dfrac{\binom{\samples}{v_{ij}}\qij{v_{ij}}(1-\qij{})^{\samples-v_{ij}}\binom{\samples}{v_{ij}}
 (\qij{'})^{v_{ij}}(1-\qij{'})^{\samples-v_{ij}} }{\binom{\samples}{v_{ij}}(\half)^\samples}\right) -1.    
\end{align}
where $V \in \mathbb{R}^{\items(\items-1)}$ is the set of all possible vectors with each element belonging to the set $\{0,1,\cdots,k\}$. There are $(k+1)^{d(d-1)}$ such vectors. We further simplify the summation over $V$ as 
\begin{align}
   \chisq(\Pbb_0 ,\P_{\mset}) = \E_{(\popq,\popq')\sim \textnormal{Unif}(\mset)} \prod_{i=1}^\items\prod_{j=1}^\items  \left(\sum_{\ell=0}^\samples\dfrac{  \binom{\samples}{\ell} \qij{\ell}(1-\qij{})^{\samples-\ell}(\qij{'})^{\ell}(1-\qij{'})^{\samples-\ell}}{(\half)^\samples }\right)- 1.
    \label{eq:chisquared}
\end{align}
This gives us the $\chisq$ divergence for the construction defined in terms of the elements of the matrices in the set $\mset$. Later, we will see that the set $\mset$ designed for the different modeling assumptions considered (namely, MST and parameter-based) consist solely of matrices with entries from the set $\{\half - \lbdist, \half, \half + \lbdist\}$. This information enables us to further simplify the expression for $\chisq(\P_0, \P_{\mset})$. \\

Consider a pair of matrices $(\popq, \popq')$ sampled uniformly at random from the set $\mset$. Let an agreement be defined as the occurrence of $\half + \lbdist$ (or $\half-\lbdist$) in the same position in $\popq$ and $\popq'$ and a disagreement is defined as the occurrence of $\half + \lbdist$ in $\popq$ or $\popq'$ in the same position as $\half -\lbdist$ in $\popq'$ or $\popq$ respectively. Next, we define two statistics $\agree$ and $\disagree$ that quantify the number of agreements and disagreements, respectively, in the matrix pair $(\popq, \popq')$ as shown here 
\begin{align}
\begin{split}
        \agree(\popq, \popq') &=  \sum_{i=1}^\items\sum_{j=1}^\items \left[\mathbbm{1}_{\{\qij{}=\qij{'}=\half+\lbdist\}} + \mathbbm{1}_{\{\qij{}=\qij{'}=\half-\lbdist\}} \right], \\
    \disagree(\popq, \popq') &= \sum_{i=1}^\items\sum_{j=1}^\items \left[\mathbbm{1}_{\{\qij{}=\half+\lbdist\}}\mathbbm{1}_{\{\qij{'}=\half-\lbdist\}} + \mathbbm{1}_{\{\qij{}=\half-\lbdist\}}\mathbbm{1}_{\{\qij{'}=\half+\lbdist\}} \right]. 
    \label{eq:agreement}
    \end{split}
\end{align}

Using these definitions, we state the following Lemma to analyse   $\chisq(\P_0, \P_{\mset})$ in \eqref{eq:chisquared}. 
\begin{lemma}
Consider two pairwise-comparison probability matrices $\popq$ and  $\popq'$ with $\qij{} \in \{\half-\lbdist, \half, \half+\lbdist\}$ and $\qij{'} \in \{\half-\lbdist, \half, \half+\lbdist\}$. Suppose $\agree(\popq, \popq') = \agree$ and $\disagree(\popq, \popq') = \disagree$. Then, we have 
\begin{align}
 \prod_{i=1}^\items\prod_{j=1}^\items  \left(\sum_{\ell=0}^\samples\dfrac{  \binom{\samples}{\ell} \qij{\ell}(1-\qij{})^{\samples-\ell}(\qij{'})^{\ell}(1-\qij{'})^{\samples-\ell}}{(\half)^\samples }\right) \leq ( 1+4\lbdist^2)^{\samples(\agree-\disagree)}. 
\end{align}
\label{lem:agreement}
\end{lemma}
The proof is provided at the end of this subsection. Using Lemma~\ref{lem:agreement} and \eqref{eq:chisquared}, we get an upper bound on the $\chisq$ divergence as 
\begin{align}
      \chisq(\Pbb_0 ,\P_{\mset}) 
  \leq \E_{(\popq,\popq')\sim \textnormal{Unif}(\mset)}\left[( 1+4\lbdist^2)^{\samples(\agree-\disagree)}\right] - 1.
\label{eq:final_chisq}
\end{align}

We conclude the background section on the converse results here. We will use the equations discussed in this section to derive the lower bound for the different modeling assumptions in Theorem~\ref{thm:lbmst} and Theorem~\ref{thm:lbbtl} in their respective proofs. 

\paragraph{Proof of Lemma~\ref{lem:agreement}}

Let 
\begin{align}
   G(\popq, \popq') = \prod_{i=1}^\items\prod_{j=1}^\items  \left(\sum_{\ell=0}^\samples\dfrac{  \binom{\samples}{\ell} \qij{\ell}(1-\qij{})^{\samples-\ell}(\qij{'})^{\ell}(1-\qij{'})^{\samples-\ell}}{(\half)^\samples }\right)- 1.
\end{align}
Let $\fung(\qij{}, \qij{'})$ denote the summation in the equation above, that is 
\begin{align}
    \fung(\qij{}, \qij{'}) = 2^\samples \sum_{\ell=0}^\samples  \binom{\samples}{\ell} \qij{\ell}(1-\qij{})^{\samples-\ell}(\qij{'})^{\ell}(1-\qij{'})^{\samples-\ell}.
    \label{eq:fqij}
\end{align}
Notice that if $\qij{} = \half $ or $\qij{'} = \half$ then $\fung(\qij{}, \qij{'}) = 1$. Additionally, $\fung(\half + \lbdist, \half + \lbdist) = \fung(\half - \lbdist, \half-\lbdist) $ and
\begin{align}
    \nonumber\fung\left(\half + \lbdist, \half + \lbdist\right) =&  2^\samples \sum_{\ell=0}^\samples \left(\half + \lbdist \right)^{2\ell} \sum_{\ell=0}^\samples \left(\half - \lbdist \right)^{2\samples - 2\ell}
    \\\nonumber =& 2^\samples \left(\half + 2\lbdist^2\right)^\samples \sum_{\ell=0}^\samples \binom{\samples}{\ell} \left(\half + \dfrac{\lbdist}{\half + 2\lbdist^2} \right)^\ell \left(\half + \dfrac{\lbdist}{\half - 2\lbdist^2} \right)^{\samples -\ell} \\
    = & (1 + 4\lbdist^2)^\samples. 
\label{eq:f1}
\end{align}
Also, note that $\fung(\half + \lbdist, \half - \lbdist) = \fung(\half - \lbdist, \half + \lbdist)$ and
\begin{align}
\nonumber\fung(\half - \lbdist, \half + \lbdist) &= 2^\samples \sum_{\ell=0}^\samples  \binom{\samples}{\ell} (\half-\lbdist)^{\ell}(\half + \lbdist)^{\samples-\ell}(\half + \lbdist)^{\ell}(\half - \lbdist)^{\samples-\ell} \\
\nonumber & = (\frac{1}{2}-2\lbdist^2)^\samples \sum_{\ell=0}^\samples  \binom{\samples}{\ell}  \\
& = (1-4\lbdist^2)^\samples.
\label{eq:f2}
\end{align}
Therefore, if the pair of matrices $\popq, \popq'$ have $\agree$ agreements and $\disagree$ disagreements, then using \eqref{eq:f1}, \eqref{eq:f2} we get 
\begin{align*}
    G(\popq, \popq') =& \fung\left(\half + \lbdist, \half + \lbdist\right)^{\agree}\fung\left(\half + \lbdist, \half - \lbdist\right)^{\disagree}  \\
    &=  (1+4\lbdist^2)^{\samples\agree} (1-4\lbdist^2)^{\samples \disagree}\\
    &\leq (1+4\lbdist^2)^{\samples(\agree-\disagree)} . 
\end{align*}
This proves Lemma~\ref{lem:agreement}. 
\subsubsection{Proof of Proposition~\ref{remarkkone}}
\label{sec:impossibility}

In this section, we provide a construction of the null and the alternate in \eqref{eq:test} under the model-free assumption that proves the statement of Proposition~\ref{remarkkone}. To this end, let $\popp_0$ be a pairwise probability matrix with the $(i,j)^{th}$ element denoted by $\pij{}$ for all $i,j \in [\items]$. We will provide more details about $\popp_0$ subsequently. Consider the case where under the null $\popp = \popq = \popp_0$ and under the alternate $\popp = \popp_0$ and $\popq \sim \textnormal{Bernoulli}(\popp_0)$. Under this notation, we have that under the alternate $\qij{}\sim\textnormal{Bernoulli}(\pij{})$.  We choose $\popp_0$ such that for all $i,j\in [\items]$ we have $0 \leq \pij{}\leq \half$. With this, we argue that under the alternate construction, for any realization of $\popq$, we have
$$
\frobnorm{\popp_0 - \popq}^2 \geq \sum_{i=1}^\items\sum_{j=1}^\items \pij{2}.
$$
In this manner, by choosing an appropriate $\popp_0$, we construct the alternate for any given $\distance$ which satisfy the conditions in the two-sample testing problem in \eqref{eq:test}. Note that the maximum value of $\frobnorm{\popp_0-\popq}^2$ attainable in this construction is when $\pij{}= \half $ for all $(i,j) \in [\items]$. In this setting, $\frobnorm{\popp_0-\popq}^2 = \frac{\items^2}{4}$, for all realizations of $\popq$. Thus, in our construction, the parameter $\distance$ is at most $\frac{1}{2}$. 

Now, in Proposition~\ref{remarkkone}, we consider the case where we have one pairwise-comparison for each pair in each population, that is, $\kpij = \kqij = 1\; \forall \; i,j \in [\items]$. Recall that the observed matrices corresponding to the two populations are denoted by $\matx$ and $\maty$ which are distributed as $\matx \sim \textnormal{Bernoulli}(\popp)$ and $\maty\sim \textnormal{Bernoulli}(\popq)$. Now, for our construction, we see that $\matx$ is distributed identically under the null and the alternate as $\matx \sim \textnormal{Bernoulli}(\popp_0)$, and $\maty$ is distributed as $\maty \sim \textnormal{Bernoulli}(\popp_0)$ under the null and $\maty \sim \textnormal{Bernoulli}(\popq)$ under the alternate. Thus, to distinguish between the null and the alternate, we must be able to distinguish between the product distribution $\Pbb_0 \coloneqq \textnormal{Bernoulli}(\popp_0)\times  \textnormal{Bernoulli}(\popp_0)$ and the product distribution $\Pbb_1 \coloneqq  \textnormal{Bernoulli}(\popp_0)\times  \textnormal{Bernoulli}(\popq)$ where $\popq \sim \textnormal{Bernoulli}(\popp_0)$.

For the setting with one comparison per pair, we have access to only first order statistics for matrices $\matx$ and $\maty$. Since the Bernoulli parameters for all pairs $(i,j)$ are independently chosen under the model-free setting, we look at the first order statistics of any pair $(i,j)$, which are given by  $\pr(\matx_{ij}=1), \pr(\maty_{ij}=1), \pr(\matx_{ij}=1,\maty_{ij}=1)$. Now, observe that under both the distributions $\Pbb_0$ and $\Pbb_1$ we have that $$
\pr(\matx_{ij}=1) = \pij{}; \; \; {}, \pr(\maty_{ij}=1)=\pij{}; \; \; \pr(\matx_{ij}=1,\maty_{ij}=1)=\pij{2}.
$$
Since the first order statistics under both distributions $\Pbb_0$ and $\Pbb_1$ are identical, we conclude that no algorithm can distinguish between these distributions with a probability of error less than half. In turn, the minimax risk defined in \eqref{eq:risk} is at least half. This proves Proposition~\ref{remarkkone}.

\subsubsection{Proof of Theorem~\ref{thm:lbmst}}
\label{sec:prooflbmst}
In this section, we establish a lower bound on the critical radius \eqref{eq:criticalrad} for the two-sample testing problem defined in \eqref{eq:test} under the assumption of the MST class as stated in Theorem~\ref{thm:lbmst}. First, we provide a construction for the null and the alternate in Section~\ref{sec:preliminaries}. In this construction, we set $\popp = \popq = \popp_0 $ under the null and $\popp=\popp_0, \popq \sim \textnormal{Unif}(\mset)$ under the alternate where $\mset$ is a set of matrices belonging to the MST class. To complete the description of the construction, we now describe the set $\mset$ for the MST class of pairwise-comparison probability matrices. The probability matrices in $\mset$ correspond to a fixed ranking of items. Each matrix in $\mset$ is such that the upper right quadrant has exactly one element in each row and each column equal to $\half + \lbdist$ for some $\lbdist \in (0,\half)$. The rest of the elements above the diagonal are half. The elements below the diagonal follow from the shifted-skew-symmetry condition imposed on MST probability matrices. It can be verified that all matrices $\popq \in \mset$ lie in the MST class. Note that the set $\mset$ has $(d/2)!$ matrices. Since each matrix has a total of $\items$ elements equal to $\half \pm \lbdist$, we get that $\frac{1}{\items^2}\frobnorm{\popp_0-\popq}^2 = \distance^2 = \frac{\lbdist^2}{\items}$. This implies $\distance^2 \leq \frac{1}{4\items}$. \\

Now, to derive bounds on the minimax risk according to \eqref{eq:mmrisk}, we analyse the $\chisq$ divergence between $\Pbb_0$ and $\Pbb_{\mset}$. From the analysis of $\chisq(\P_0, \P_{\mset})$ in Section~\ref{sec:preliminaries}, specifically \eqref{eq:final_chisq}, we have that 
 \begin{align*}
     \chisq(\Pbb_0 ,\P_{\mset}) &\leq \E_{(\popq,\popq')\sim \textnormal{Unif}(\mset)}\left[( 1+4\lbdist^2)^{\samples(\agree-\disagree)}\right] - 1
 \end{align*}
where $\agree$ and $\disagree$ are the number of agreements and disagreements between the matrices $\popq, \popq'$, as defined in \eqref{eq:agreement}. 
\noindent Now, to compute the $\chisq$ divergence, we want to find the probability that two matrices picked uniformly at random from $\mset$ have $i$ agreements in the upper right quadrant(the total number of agreements is $2i$ due to shifted-skew-symmetry). Given a matrix $\popq$ from set $\mset$, for $i$ agreements, we want to choose a matrix $\popq' \in \mset$ such that exactly $i$ of the perturbed elements share the same position as in $\popq$. There are $\binom{\items/2}{i}$ ways of choosing the $i$ elements. Now that the $i$ elements have their position fixed, we have to find the number of ways we can rearrange the remaining $\frac{\items}{2} - i$ elements such that none of them share a position with the remaining perturbed elements in $\popq$. This problem is the same as reshuffling and matching envelopes with letters such that no letter matches with originally intended envelope. The number of ways to rearrange a set of $i$ objects in such a manner is given by $i!(\frac{1}{2!}-\frac{1}{3!} + \cdots + (-1)^i\frac{1}{i!})$. Thus the number of ways of rearrangement for $\frac{\items}{2} - i$ items is upper bounded by $\half(\items/2-i)!$. Thus, the probability of $2i$ agreements is upper bounded as 
\begin{align}
\Pbb(\agree = 2i) \leq \dfrac{(\items/2)!}{(\items/2-i)!i!}\dfrac{(\items/2-i)!}{2(\items/2)!} \leq \dfrac{1}{2(i)!}.    
\label{eq:prob-bound}
\end{align}
Then, we further simplify \eqref{eq:final_chisq} as
\begin{align}
\nonumber\chisq(\Pbb_0 ,\P_{\mset}) &\leq \E_{(\popq,\popq')\sim \textnormal{Unif}(\mset)}\left[( 1+4\lbdist^2)^{\samples(\agree-\disagree)}\right] - 1\\
 \nonumber   &\leq \E_{(\popq,\popq')\sim \textnormal{Unif}(\mset)}\left[( 1+4\lbdist^2)^{\samples\agree}\right] -1\\
    &\leq \sum_{i=0}^{\items/2}\Pbb(\agree = 2i)(1+4\eta^2)^{2ki} -1.
    \label{eq:mstchi}
\end{align}
Notice that if we choose $\samples = \dfrac{\const}{4\lbdist^2}$ with some constant $\const \in (0,1)$ then we have that 
\begin{align*}
    (1+4\lbdist^2)^\samples &= \sum_{\ell=0}^\samples \binom{\samples}{\ell}(4\lbdist^2)^\ell\\
    & \leq 1 + \sum_{\ell=1}^\samples(4\lbdist^2\samples)^\ell\\
    &\leq 1 + \sum_{\ell=1}^\samples \const^\ell \\
    &\leq 1 + \const',
\end{align*}
where $\const'$ is some positive constant. Using this, we show that the $\chisq$ divergence in \eqref{eq:mstchi} is upper bounded by a constant for $\samples \leq \dfrac{\const}{\lbdist^2}$, as follows 
\begin{align*}
     \chisq(\Pbb_0 ,\P_{\mset})  &\leq \sum_{i=0}^{\items/2} \Pbb(\agree = 2i)(1+4\lbdist^2)^{2\samples i} - 1 \\
     &\leq \sum_{i=0}^{\items/2} \Pbb(\agree = 2i)(1+\const')^{2i} -1 \\
     &=\sum_{i=0}^{\items/2}\Pbb(\agree = 2i) - 1 + \sum_{i=0}^{\items/2}\Pbb(\agree = 2i)((1+\const')^{2i} - 1)\\
     &\stackrel{(a)}{\leq} \sum_{i=0}^{\items/2} \dfrac{1}{2(i!)}((1+\const')^{2i}-1)\\
     &\leq \half\left(\exp((1+\const')^2)-\exp(1) + \sum_{i=\items/2}^{\infty}\dfrac{1}{i!}\right)\\
     &\leq \const'',
\end{align*}
where $\const''$ is some positive constant. The inequality (a) follows from \eqref{eq:prob-bound}.
This proves that there exists a constant $\const$, such that if $\samples \leq \dfrac{\const}{\lbdist^2} = \dfrac{\const}{\items\distance^2}$, then the $\chisq$ divergence is upper bounded by $\frac{1}{9}$. According to \eqref{eq:mmrisk}, this implies that the minimax risk is at least $\frac{1}{3}$.
This establishes the lower bound on the critical testing radius for two-sample testing under the MST modeling assumption as $\distance^2_\modelclass > \const\dfrac{1}{\samples\items}$ and proves Theorem~\ref{thm:lbmst}.

\subsubsection{Proof of Theorem~\ref{thm:lbbtl}}
\label{sec:prooflbbtl}

Consider any arbitrary non-decreasing function $\fun : \R \rightarrow [0,1]$ such that $\fun(\thetaa) = 1 - \fun(-\thetaa)~~~\forall\ \thetaa\in\R$. In order to prove the lower bound on testing radius stated in Theorem~\ref{thm:lbbtl}, we construct a set of matrices $\mset$ based on the parameter-based pairwise-comparison probability model described in \eqref{eq:parametric} associated to the given function $\fun$. Observe that $\fun(0) = \half$. Recall that under the parameter-based model the sum of all weights is fixed as $\sum_{i=1}^\items \weight_i = 0$. We use the weight parameter to define the construction for the lower bound. 

Recall the null and alternate construction described in Section~\ref{sec:preliminaries} to prove the lower bound. Accordingly, we set $\popp = \popq = \popp_0 $ under the null and $\popp=\popp_0, \popq \sim \textnormal{Unif}(\mset)$ under the alternate where $\mset$ is a set of matrices belonging to the parameter-based class. The weights $\weight_{\popp_0} = [0,\cdots, 0]\in \mathbb{R}^\items$ correspond to the pairwise probability matrix $\popp_0 = [\half]^{\items \times \items }$. Now for creating the set $\mset$, consider a collection of weight vectors $\weight_{\mset}$ each with half the entries as $ \perturb$ and the other half as $- \perturb$, thereby ensuring that $\sum_{i \in [\items]}\weight_i = 0$.  We set $\perturb$ to ensure that each of the probability matrices induced by this collection of vectors obey $\frac{1}{\items}\frobnorm{\popp_0 - \popq}= \distance$. We define the set of matrices $\mset$ as the set of pairwise-comparison probability matrices induced by the collection of values of $\weight_{\mset}$. Clearly, there are $\binom{\items}{\items/2}$ matrices in $\mset$. A similar argument holds for odd $\items$ wherein $\frac{\items-1}{2}$ elements of the weight vector are $\perturb$ and $\frac{\items-1}{2}$ elements are $ -\perturb$. 
Since $f$ is monotonic, $\fun(-2\perturb) \le \fun(0) \le \fun(2\perturb)$ and we have that $\fun(2\perturb)  = 1- \fun(-2\perturb)$, we define $\fun(-2\perturb)  = \half - \lbdist $ and $\fun(2\perturb) = \half + \lbdist$ for some $0 < \lbdist \le \half$. \\

Similar to the proof of Theorem~\ref{thm:lbmst}, we use \eqref{eq:final_chisq} to bound the $\chisq$ divergence between the null and the alternate constructed. We first note that if we sample two matrices $\popq$ and $\popq'$ uniformly at random (with replacement) from $\mset$, then if the number of agreements is $\frac{i^2}{2}$ then the number of disagreements is equal to $\frac{(\items-i)^2}{2}$. The probability of $\frac{i^2}{2}$ agreements is given by
\begin{align*}
    \Pbb\left(\agree = \frac{i^2}{2}, \disagree = \frac{(\items-i)^2}{2}\right) =  \dfrac{\binom{\items/2}{i/2}\binom{\items/2}{i/2}}{\binom{\items}{\items/2}}.
\end{align*}
Following from \eqref{eq:chisquared}, the $\chisq$ divergence is 
\begin{align*}
\chisq(\Pbb_0 ,\P_{\mset}) &= \E_{(\popq,\popq')\sim \textnormal{Unif}(\mset)}\left[( 1+4\lbdist^2)^{\samples(\agree-\disagree)}\right] - 1\\
    &\leq \sum_{i=0}^d \dfrac{\binom{\items/2}{i/2}\binom{\items/2}{i/2}}{\binom{\items}{\items/2}} (1+4\lbdist^2)^{k(i^2-(\items-i)^2)/2} - 1.
\end{align*}
For ease of presentation, we replace $\frac{\items}{2}$ by $\dbytwo$ and $\frac{i}{2}$ by $\ibytwo$, to get 
\begin{align}
    \nonumber \chisq(\Pbb_0 ,\P_{\mset}) &\leq \sum_{\ibytwo=0}^\dbytwo \dfrac{\binom{\dbytwo}{\ibytwo}\binom{\dbytwo}{\ibytwo}}{\binom{2\dbytwo}{\dbytwo}} (1+4\lbdist^2)^{2\samples(2\ibytwo\dbytwo - \dbytwo^2)} - 1\\
    \nonumber &\leq   \sum_{\ibytwo=0}^\dbytwo  \left(\half\right)^\dbytwo \binom{\dbytwo}{\ibytwo}(1+ 4\lbdist^2)^{2\samples(2\ibytwo\dbytwo - \dbytwo^2)} - 1\\
    \nonumber &\leq   \sum_{\ibytwo=0}^\dbytwo  \left(\half\right)^\dbytwo \binom{\dbytwo}{\ibytwo}\exp({8\lbdist^2\samples(2\ibytwo\dbytwo - \dbytwo^2)} )- 1.
    \end{align}
Here, we see that the summation in the final expression is equal to the expectation of $\exp({8\lbdist^2\samples(2\ibytwo\dbytwo - \dbytwo^2)} )$ over the random variable $\ibytwo$ where $\ibytwo \sim \Bin(\dbytwo, \half)$. So, 
\begin{align}
   \nonumber \chisq(\Pbb_0 ,\P_{\mset})  &\leq  \E_\ibytwo\left[ \exp({8\lbdist^2\samples(2\ibytwo\dbytwo - \dbytwo^2)} )\right] -  1 \\
      &\leq  \sum_{i=0}^\infty \dfrac{(8\lbdist^2\samples\dbytwo)^i}{i!}\E_\ibytwo\left[(2\ibytwo-\dbytwo)^{i}\right] - 1  ,
     \label{eq:inequalitybtl}
\end{align}
where $\E_\ibytwo[(2\ibytwo-\dbytwo)^i]$ is the scaled centered $i^{\textnormal{th}}$ moment of $\Bin(\dbytwo, \half)$. To get the expression for the centered moments, we first find the moment generating function of the random variable $\ibytwo' = 2\ibytwo -\dbytwo$, as
\begin{align*}
    \Ebb\left[e^{(2\ibytwo-\dbytwo)t}\right] &= e^{-\dbytwo t}\Ebb\left[e^{2\ibytwo t}\right]\\
    &= e^{-\dbytwo t}\sum_{\ibytwo=0}^\dbytwo \binom{\dbytwo}{\ibytwo}\left(\half e^{2t}\right)^\ibytwo \left(\half\right)^{\dbytwo-\ibytwo}\\
    & = e^{-\dbytwo t}\left(\half + \half e^{2t}\right)^\dbytwo\\
    &= \left(\dfrac{e^{-t} + e^{t}}{2}\right)^{\dbytwo}\\
    & = (\cosh{t})^\dbytwo.
\end{align*}
 Then, we have 
 \begin{align*}
     \Ebb \left[(2\ibytwo-\dbytwo)^i\right] = \dfrac{d^i (\cosh{t})^\dbytwo}{dt^i} \bigg|_{t=0}, 
 \end{align*}
 which is the $i^{\textnormal{th}}$ derivative of $(\cosh{t})^\dbytwo$ evaluated at $t = 0$. 
 This leads to the fact that for odd $i$,  $\E[(2\ibytwo-\dbytwo)^i] = 0$ and for even $i$, $\E[(2\ibytwo-\dbytwo)^i] \leq \frac{i!}{(i/2)!}\dbytwo^{i/2}$. Using this with \eqref{eq:inequalitybtl}, we get 
\begin{align}
     \chisq(\Pbb_0 ,\P_{\mset})  &\leq\sum_{i=1}^\infty \const_i(64\lbdist^4\samples^2\dbytwo^{3})^i
\end{align}
where $\const_i = \frac{1}{(i/2)!}$, that is, $\const_i$ is decreasing as $i$ increases.

Thus, we see that if $\samples \leq \dfrac{\const}{\lbdist^2\items^{3/2}}$ then there is a small enough $\const$ such that the $\chisq$ divergence is upper bounded by $\frac{1}{9}$. In this construction, we have $\distance^2 =\lbdist^2/2$, therefore, using \eqref{eq:mmrisk}, the lower bound for two-sample testing under the parameter-based modeling assumption is given as $\distance^2 = \Omega\left(\dfrac{1}{\samples\items^{3/2}}\right)$. This proves Theorem~\ref{thm:lbbtl}.

\subsubsection{Proof of Theorem~\ref{thm:polytimelb}}
 \label{sec:proofpolytime}
 To prove Theorem~\ref{thm:polytimelb}, we use the conjectured average-case hardness of the planted clique problem. In informal terms, the planted clique conjecture asserts that it is hard to detect the presence of a planted clique in an Erd\" os-R\'enyi random graph. In order to state it more precisely, let $G(\items, \pcsize)$ be a random graph on $\items$ vertices constructed in one of the following two ways:

$H_0$ : Every edge is included in $G(\items, \pcsize)$ independently with probability $\half$

$H_1$ : Every edge is included in $G(\items, \pcsize)$ independently with probability $\half$. In addition, a set of $\pcsize$ vertices is chosen uniformly at random and all edges with both endpoints in the chosen set are added to $G$.

The planted clique conjecture then asserts that when $\pcsize = o(\sqrt{\items})$, then there is no polynomial-time algorithm that can correctly distinguish between $H_0$ and $H_1$ with an error probability that is strictly bounded below $\half$. We complete the proof by identifying a subclass of SST matrices and showing that any testing algorithm that can distinguish between the subclass of SST matrices and the all half matrix, can also be used to detect a planted clique in an Erd\" os-R\'enyi random graph.

Consider the null with $\popp=\popq = [\half]^{\items \times \items}$ and the alternate such that $\popp = [\half]^{\items\times\items}$ and $\popq$ is chosen uniformly at random from set $\mset$. The set of probability matrices $\mset$ contains all $(\items\times\items)$ matrices with the upper left and lower right quadrant equal to all half, the upper right quadrant is all half except a $(\pcsize\times\pcsize)$ planted clique (i.e., a $(\pcsize,\pcsize)$ submatrix with all entries equal to one). 
Then we have $\distance^2 = \pcsize^2/2\items^2$. The bottom left quadrant follows from the skew symmetry property. Recall that we observe one sample per pair of items $(i > j)$. This testing problem is reduced to a goodness-of-fit testing problem as shown in \eqref{eq:goodnessoffit} and \eqref{eq:goodnessoffit_test}.

Consider the set of $\left(\frac{\items}{2} \times \frac{\items}{2}\right)$ matrices comprising the top right $\left(\frac{\items}{2} \times \frac{\items}{2}\right)$ sub-matrix of every matrix in $\mset$. We claim that this set is identical to the set of all possible matrices in the planted clique problem with $\frac{\items}{2}$ vertices and a planted clique of size $ \pcsize$. Indeed, the null contains the all-half matrix corresponding to the absence of a planted clique, and the alternate contains all symmetric matrices that have all entries equal to half except for a $\left(\pcsize, \pcsize\right)$ all-ones sub-matrix corresponding to the planted clique. We choose the parameter $\kappa = \frac{\sqrt{\items}}{\log\log(\items)}$ so that any constant multiple of it will be within the hardness regime of planted clique (for sufficiently large values of $\items$). Now, we leverage the planted clique hardness conjecture to state that the null in our construction cannot be distinguished from the alternate by any polynomial-time algorithm with probability of error less than $\half$. This implies that for polynomial-time testing it is necessary that $\distance^2 \geq \dfrac{\const}{\items(\log\log(\items))^2}$. This proves Theorem~\ref{thm:polytimelb}.

\subsection{Proof of Theorem~\ref{thm:ell_length}}
\label{sec:proofalgo1}\paragraph{Bounding the Type I error.}
In this proof, we first bound the Type I error and subsequently bound the Type II error. To bound the probability of error of Algorithm~\ref{testalgo_partial}, we study the distribution of the test statistic $\stat$ under the null and the alternate. Algorithm~\ref{testalgo_partial} uses the test statistic defined in \eqref{eq:teststat}. To understand the distribution of the test statistic, we first look at the distribution of $\matx_{ij}$ and $\maty_{ij}$. 
 
 In Algorithm~\ref{testalgo_partial}, we break the partial ranks into disjoint pairwise-comparisons. Under the Plackett-Luce model disjoint pairwise-comparisons obtained from the same partial ranking are mutually independent. Additionally, since the Plackett-Luce model obeys the property of independence of irrelevant alternatives, the probability of observing item $i$ being ranked ahead of item $j$ in a partial ranking is independent of the other items being ranked in that partial ranking. Thus, for any pair of items $(i,j)$, the probability of $i$ beating $j$, conditioned on the event that the pair $(i,j)$ was observed, is always equal to $\pij{}$ for the population corresponding to pairwise probability matrix $\popp$ and $\qij{}$ for the population corresponding to pairwise probability matrix $\popq$. This holds true irrespective of which other items are involved in that partial (or total) ranking. With this in mind, we identify the distribution of $\matx_{ij}$ conditioned on $\kpij$  as $\matx_{ij}\,|\,\kpij\sim \Bin(\kpij, \pij{}) $. Similarly, we have $\maty_{ij}\,|\,\kqij \sim \Bin(\kqij, \qij{})$. 
  Let $\textbf{\samples}^\distra, \textbf{\samples}^\distrb$ denote the vector of $\kpij, \kqij$ for all $(i,j), \, i<j$. The conditional expectation of $\stat$ is 
\[
\Ebb[\stat\,|\, \textbf{\samples}^\distra,\textbf{\samples}^\distrb] = \sum_{i=1}^{j-1}\sum_{j=1}^\items \dfrac{\mathbb{I}_{ij}\kpij\kqij}{\kpij+\kqij}(\pij{}-\qij{})^2     .
\]
Under the null we have $\pij{}= \qij{}$. Clearly, using the law of total expectation, we see that $\E_{\nullh}[\stat] = \E[\;\E[\stat\,|\,\textbf{\samples}^\distra, \textbf{\samples}^\distrb]\;] = 0$. We now upper bound the variance of $\stat$ under the null. Recall from \eqref{eq:lawtotalvar} and \eqref{eq:var1} that 
\begin{align}
\nonumber    \Var_{\nullh}[\stat] &\leq \sum_{i=1}^{j-1}\sum_{j=1}^\items 8\mathbb{I}_{ij}\left(\dfrac{ \kpij\kqij}{(\kpij+\kqij)}(\pij{}-\qij{})^2 + 3\right) + \Var[\,\E[\stat|\textbf{\samples}^\distra, \textbf{\samples}^\distrb]\,]\\
    &\leq 24\items^2.
    \label{eq:varbound}
\end{align}
Now, we have the information to bound the Type I error. To get a bound on the Type I error with threshold $\threshold$, we use the one sided Chebyshev's inequality,
\begin{align}
    \Pbb_{\nullh}(\stat \geq \threshold) \leq \dfrac{\Var_{\nullh}[\stat]}{\Var_{\nullh}[\stat] + \threshold^2}.
\label{eq:onesidedchebyshev}
\end{align}
Using the bound in \eqref{eq:varbound} and \eqref{eq:onesidedchebyshev}, we observe that if $\threshold = 11\items$ then the Type I error is at most $\frac{1}{6}$.
This concludes the proof that Algorithm~\ref{testalgo_partial} controls the Type I error of the test \eqref{eq:test_pl} at level $\frac{1}{6}$. 

\paragraph{Bounding the Type 2 error.} We now analyse the Type II error of Algorithm~\ref{testalgo_partial}, that is, the probability of our algorithm failing to reject the null, under the alternate. We consider two cases depending on whether the pairwise-comparison data created through the rank breaking method has at least $\samples$ pairwise-comparisons per pair $(i,j),\, i<j$, or not, for some positive integer $\samples$. We will define $\samples$ later in the proof. Let the case where the pairwise-comparisons created in each population have at least $\samples$ comparisons of each pair be denoted by $\case$ and let the associated Type II error be denoted by $\typell_1$. Let the Type II error associated with the remaining case be denoted by $\typell_2$. Our objective is to provide an upper bound on the total Type II error which is  $\typell  = \Pbb(\case)\typell_1 + (1-\Pbb(\case))\typell_2$.

First, we derive a bound on $\Pbb(\case)$. To start, we note that the probability of observing a specific pair from a total ranking is $\frac{1}{\items}$ if $\items$ is odd and $\frac{1}{\items-1}$ if $\items$ is even. Recall that for a given $\length$, each sample is a ranking of some $\length$ items chosen uniformly at random from the set of $\items$ items. Under this setting, we see that the probability that ``Random disjoint'' rank breaking yields a specific pairwise-comparison from a $\length$-length partial ranking is $\frac{\length}{\items(\items-1)}$ if $\length$ is even and $\frac{\length-1}{\items(\items-1)}$ if $\length$ is odd. Henceforth, in this proof, we assume that $\length$ is even. The proof follows similarly for odd $\length$. Thus, the number of pairwise-comparisons observed of any pair $(i,j)$ is a binomial random variable with Bernoulli parameter $\frac{\length}{\items(\items-1)}$. Consequently, if we have $\numranks$ samples from each population, then for the population corresponding to the pairwise probability matrix $\popp$, for all pairs $(i,j)$ we have $\kpij \sim \Bin(\numranks, \frac{\length}{\items(\items-1)})$. Similarly for the population corresponding to pairwise probability matrix $\popq$, for all pairs $(i,j)$ we have $\kqij\sim\Bin(\numranks, \frac{\length}{\items(\items-1)})$. Now, we are equipped to compute the probability of case $\case$. We divide the samples available in each population into $\samples$ sections of equal sizes. Let the samples in each population be indexed from $1$ to $\numranks$ then we assign the first $\lfloor \frac{\numranks}{\samples}\rfloor$ into the first section and so on. Now, we know that the probability of observing a pair $(i,j)$ at least once in one such section is given by $1 - (1-\frac{\length}{\items(\items-1)})^{\lfloor \frac{\numranks}{\samples}\rfloor}$. Using this, we get  the following union bound,  
\begin{align*}
    \Pbb(\kpij \geq \samples) \geq   1 - \samples\left(1-\frac{\length}{\items(\items-1)}\right)^{\frac{\numranks}{\samples}} .
\end{align*}
The same inequality holds for $\kqij$ for all $(i,j)$. Then, the probability that all pairs of items had at least $\samples$ pairwise-comparisons in both populations is lower bounded as  
\begin{align*}
    \Pbb(\case)  &\geq 1 - \dfrac{2\samples\items(\items-1)}{2}\left(1 - \frac{\length}{\items^2}\right)^{\frac{\numranks}{\samples}} \\
    &\geq 1- \samples\items^2\exp\left(-\frac{\numranks\length}{\samples\items^2}\right)
\end{align*}
We see that, if $\numranks = \const\samples\items^2\log(\items)/\length$ for some positive constant $\const$, then $\Pbb(\case) \geq 1-\frac{\samples}{\items^{\const-2}}$.\\

\noindent Conditioned on the case $\case$, we invoke Theorem~\ref{thm:upperbound} to control the Type II error. Recall that Theorem~\ref{thm:upperbound} asserts that there is a constant $\const_0 >0$ such that if we have $\samples$ pairwise-comparisons of each pair $(i,j)$ from each population where $\samples \geq \max\{\const_0\frac{1}{\items \distance^2},2\}$ then the Type II error of Algorithm~\ref{testalgo} for the testing problem \eqref{eq:test_pl} is upper bounded by $\frac{1}{12}$. To apply this result to Algorithm~\ref{testalgo_partial} conditioned on case $\case$, we keep $\samples$ pairwise-comparisons for each pair where $\samples = 2\lceil\const_0\frac{1}{\items \distance^2}\rceil$. Observe that, since we assume $\distance \geq \const_1\items^{-\const_2}$ for some positive constants $\const_1$ and $\const_2$, we get the inequality $\samples \leq \const_1'\items^{\const_2'}$ for some positive constants $\const_1'$ and $\const_2'$. Under this inequality, we get that there exist positive constants $\const, \const_1'$ and $\const_2'$ such that $\Pbb(\case) > 11/12$. \\

\noindent Next, we observe that the Type II error conditioned on the complement of $\case$ is at most 1. Therefore, the total probability of failing to reject the null under the alternate is given by 
\begin{align*}
   \typell &= \Pbb(\case)\typell_1 + (1-\Pbb(\case))\typell_2 \\
   &\leq \frac{1}{12} + \frac{1}{12} = \frac{1}{6}.
\end{align*}
This concludes the proof that for some constant $C>0$, if $\numranks \geq C\dfrac{\items^2\log(\items)}{\length}\lceil \dfrac{\const_0}{\items\distance^2}\rceil$, then the probability of error of Algorithm~\ref{testalgo_partial} is at most $\frac{1}{3}$. 

\subsection{Proof of Theorem~\ref{thm:full_length}}
\label{sec:prooffulllength}
The proof of Theorem~\ref{thm:full_length} follows similarly to the proof of Theorem~\ref{thm:ell_length}. Both theorems establish the performance of Algorithm~\ref{testalgo_partial} for the two-sample testing problem stated in \eqref{eq:test_pl}. The difference lies in the assumption on the partial ranking data available and consequently in the rank breaking algorithm used. In Theorem~\ref{thm:full_length}, we assume we have total ranking data that is then deterministically converted to pairwise-comparisons in the following manner. We have a total of $\numranks$ total rankings available from each population. We divide these into subsets each containing $\items$ rankings as described in the ``Deterministic disjoint'' rank breaking method. Notice that we can break the ranking data available in a section into pairwise-comparisons such that we observe each unique pair of items at least one time. We prove this statement at the end of the section. We repeat this breaking technique for all subsets. Consequently, we get $\samples = \lfloor \frac{\numranks}{\items}\rfloor\geq 2\lceil\const\dfrac{1}{\items\distance^2}\rceil$ pairwise-comparisons for all pairs $(i,j)$ from each population. With this in mind, we apply Theorem~\ref{thm:upperbound} to obtain the desired result. 

Finally, to complete the proof we show that it is indeed possible to break $\items$ total rankings such that we observe each of $\binom{\items}{2}$ unique pairs at least once. We use a mathematical induction based argument. As a first step we observe that our hypothesis is true for $\items = 2$. In the inductive step, we assume that the hypothesis is true for all natural numbers $\items \in \{2,\cdots, r\}$. Now, we wish to prove the hypothesis is true for $\items = r+1$. First, consider the case where $r$ is even. We divide the set of $r$ items into two groups with $r/2$ items in each. From the inductive step we know that our hypothesis is true for $\items = r/2$. Consequently, we get $2 \binom{r/2}{2}$ unique pairs from within the two groups which use $r/2$ total rankings. Next, we arrange the items in group one in a list against the items in group two and make pairs by choosing the items in $i^{th}$ position in both the lists. This gives the breaking for one total ranking. We do this $r/2$ times, each time cyclically shifting the first list by one item. This step gives $r^2/4$ unique pairs that are different from the pairs obtained in the previous step and uses $r/2$ total rankings. This proves our hypothesis for $\items = r$ for even $r$. To prove our hypothesis for odd $r$, we prove our hypothesis for $r+1$ which is even using the same method described in the previous step. We complete our proof by noting that if the hypothesis is true for even $r$ then it must be true for $r-1$. This concludes our proof of Theorem~\ref{thm:full_length}
 
\subsection{Proof of Theorem~\ref{thm:permutation_test}}
\label{sec:proofalgo2}

The idea of the proof is to show that under the null hypothesis, a ranking sample sourced from the first population is mutually independent of and identically distributed as a ranking sample sourced from the second population. If this statement is true, then shuffling the population labels of ranking data, does not alter the distribution of the test statistic \eqref{eq:teststat}. This in turn controls the Type I error of the permutation test method. 

 Under the null, for some distribution $\rankdist{}$ over all total rankings, we have that $\rankdist{\popp} = \rankdist{\popq} = \rankdist{}$. This implies that under the marginal probability based model, the probability of any given partial ranking over a set of items is the same for both the populations. Specifically, conditioned on the set of items being ranked, each partial ranking in each population is sampled independently and identically, according to the distribution $\rankdist{}$. Recall that the set of items being ranked in each population is sampled independently and identically from some distribution over all non-empty subsets in $[\items]$. Consequently, each ranking sample is independent of all other ranking samples obtained from the two populations. Moreover, using the law of total probability over all the non-empty subsets in $[\items]$, we get that each ranking sample obtained in each population is identically distributed. With this, we conclude that shuffling the population labels of ranking data does not alter the distribution of the test statistic. Thus, for a permutation test with $\iterb$ iterations, the $p$-value of the test is distributed uniformly over $\{0,1/\iterb, 2/\iterb, \cdots, 1\}$. Hence, for any given significance level $\typel \in (0,1)$, by applying a threshold of $\typel$ on the $p$-value of the test, we are guaranteed to have Type I error at most $\typel$.

\section{Discussion and open problems}
\label{sec:discussion}
We conclude with a discussion focused on open problems in this area. We provide algorithms for two-sample testing on pairwise-comparison and ranking data to distinguish between two potentially different populations (in terms of their underlying distributions). Through our analysis, we see that our testing algorithm for pairwise-comparison data is simultaneously minimax optimal under the model-free setting as well as the MST and WST model. There is a gap between the testing rate of our algorithm and our information-theoretic lower bound for the SST and parameter-based models, and closing this gap is an open problem of interest.  Second, in the future, our work may help in studying two-sample testing problems pertaining to more general aspects of data from people such as function evaluations~\citep{xu2020regression,noothigattu2018loss}, issues of calibration~\cite{wang2019your}, and strategic behavior~\citep{xu2019strategyproof}. 
Thirdly, in practice we use the permutation method to calibrate our tests, ensuring valid Type I error control even when the distribution of the test statistic is difficult to characterize analytically (for instance, 
in the setting with partial rank data). Understanding the power of tests calibrated via the permutation method is an active area of research~\citep{kim2020minimax} and it would be interesting to understand this in the context of the
tests developed in our work. 
Finally, the literature on analyzing pairwise-comparison data builds heavily on probability models from social choice theory (some are described in this work). A natural related question (that has received some recent attention~\citep{seshadri2019fundamental}) is the design of goodness-of-fit hypothesis tests, that is, testing whether given pairwise-comparison data obeys certain modeling assumptions.

\section*{Acknowledgments}
This work was supported in part by NSF grants DMS 1713003, CCF 1763734, 1755656 and CAREER 1942124.

\newpage

\appendix
\section*{Appendix A. Additional details of experiments}

\noindent We now provide more details about the experiments described in Section~\ref{sec:real_data_experiment}. 

\subsection*{Ordinal versus cardinal}
\label{sec:ordinalcardinal}

The data set from~\cite{shah2016estimation} used in the ``Ordinal versus cardinal'' experiment comprises of six different experiments on Amazon Mechanical Turk crowdsourcing platform. We describe each experiment briefly here. 

\begin{itemize}[leftmargin=*]
    \item Photo age: There are 10 objects in this experiment wherein each object is a photograph of a different face. The worker is either shown pairs of photos together and asked to identify the older of the two or they provide the numeric age for each photo. There are a total of 225 ordinal responses and 275 cardinal-converted-to-ordinal responses. 
    \item Spelling mistakes: There are 8 objects in this experiment wherein each object is a paragraph of text in English possibly with some spelling mistakes. The worker is either shown pairs of paragraphs and asked to identify the paragraph with more spelling mistakes or they are asked to provide the count of spelling mistakes for all 8 paragraphs. There are a total of 184 ordinal responses and 204 cardinal-converted-to-ordinal responses. 
    \item Distances between cities: There are 16 objects in this experiment wherein each object is a pair of cities (no two objects share a common city). The worker is either shown two pairs of cities at a time and asked to identify the pair that is farther from each other, or they are asked to estimate the distances for the 16 pairs of cities. There are a total of 408 ordinal responses and 392 cardinal-converted-to-ordinal responses. 
    \item Search results: There are 20 objects in this experiment wherein each object is the result of an internet based search query of the word "internet". The worker is either asked to compare pairs of results based on their relevance or they are shown all the results and asked to rate the relevance of each result on a scale of 0-100. There are a total of 630 ordinal responses and 370 cardinal-converted-to-ordinal responses.  
    \item Taglines: There are 10 objects in this experiment wherein each object is a tagline for a product described to the worker. The worker is either asked to compare the quality of pairs of taglines or they are asked to provide ratings for each tagline on a scale of 0-10. There are a total of 305 ordinal responses and 195 cardinal-converted-to-ordinal responses.
    \item Piano : There are 10 objects in this experiment wherein each object is a sound clip of a piano key played at a certain frequency. The worker is either given pairs of sound clips and asked to identify the clip with the higher frequency or they are asked to estimate the frequency of the 10 clips. There are a total of 265 ordinal responses and 235 cardinal-converted-to-ordinal responses.
\end{itemize}
In our main experiment, we combine the data from all the experiments described above and test for statistically significant difference between the underlying distributions for ordinal responses and ordinal-converted-to-cardinal responses. We also test for difference in each individual experiment (which however have smaller sample sizes), the results are provided in Table~\ref{tab:cardinalvsordinal}. We observe that the qualitatively more subjective experiments (photo age, search results, taglines) have a lower $p$-value, which indicates that the ordinal responses are more different from cardinal-converted-to-ordinal responses in a more subjective setting.  
\begin{table}[ht!]
 \centering
 \begin{tabular}{|c|c|c|c|c|c|c|c|}
 \hline
 Experiment & Combined &  Age &Spellings& Distances&Search results & Taglines & Piano \\ \hline
 $p$-value    & 0.003 & 0.001     & 0.657             & 0.75           & 0.187            & 0.0829   & 0.514  \\ \hline
 \end{tabular}
\caption{$p$-value of two-sample test comparing the distribution of ordinal responses and the distribution of cardinal-converted-to-ordinal responses in the experiments described above }
 \label{tab:cardinalvsordinal}
 \end{table}

\subsection*{European football leagues}
\label{sec:football}
In this experiment, we obtain the match scores for four different European football leagues (English Premier League, Bundesliga, Ligue 1, La Liga) across two seasons (2016-2017, 2017-2018). There were 17 teams that played the two seasons in each of EPL, La Liga, Ligue 1 and 16 teams in Bundesliga. To test for statistically significant difference between the relative performance of the participating teams in the two consecutive seasons we combined the data from all four leagues. We also tested for difference in each individual league, the results are displayed in Table~\ref{tab:football}. From the 2016-2017 season we have 202 pairwise-comparisons in EPL, 170 pairwise-comparisons in Bundesliga, 215 pairwise-comparisons in La Liga, and 201 pairwise-comparisons in Ligue 1. From the 2017-2018 season we have 214 pairwise-comparisons in EPL, 178 pairwise-comparisons in Bundesliga, 208 pairwise-comparisons in La Liga, and 201 pairwise-comparisons in Ligue 1. From the number of comparisons available our test does not detect any significant difference between the relative performance of teams in European football leagues over two consecutive seasons. 
\begin{table}[ht!]
 \centering
 \begin{tabular}{|c|c|c|c|c|c|}
 \hline
 League &  Combined & EPL & Bundesliga & La Liga &Ligue 1  \\ \hline
 $p$-value  &  0.971 &    0.998 &     0.691       &      0.67   & 0.787    \\ \hline
 \end{tabular}
 \caption{$p$-value of two-sample test comparing relative performance of teams in a football league over two consecutive seasons. }
 \label{tab:football}
 \end{table}

\bibliographystyle{apalike}
\bibliography{bibtex}

\end{document}